\documentclass{article}

\PassOptionsToPackage{numbers,compress}{natbib}

\usepackage[main, final]{neurips_2026}% the "default" option is equal to the "main" option, which is used for the Main Track with double-blind reviewing.

\usepackage[utf8]{inputenc} % allow utf-8 input
\usepackage[T1]{fontenc}    % use 8-bit T1 fonts
\usepackage{hyperref}       % hyperlinks
\usepackage{url}            % simple URL typesetting
\usepackage{booktabs}       % professional-quality tables
\usepackage{amsfonts}       % blackboard math symbols
\usepackage{nicefrac}       % compact symbols for 1/2, etc.
\usepackage{microtype}      % microtypography
\usepackage{xcolor}         % colors
\usepackage{amsmath}
\usepackage{graphicx}
\usepackage{enumitem}
\usepackage[table]{xcolor}
\usepackage{amsmath}
\usepackage{algorithm}
\usepackage{algpseudocode}
\usepackage[dvipsnames]{xcolor}
\usepackage{makecell}
\usepackage{graphicx}
\usepackage{tabularx}
\usepackage{array}
\usepackage{makecell}
\usepackage{caption}
\usepackage{placeins}
\definecolor{bestcolor}{RGB}{244,183,181}   
\definecolor{secondcolor}{RGB}{247,210,170}  
\definecolor{thirdcolor}{RGB}{247,243,163}   
\newcommand{\best}[1]{\cellcolor{bestcolor}{#1}}
\newcommand{\second}[1]{\cellcolor{secondcolor}{#1}}
\newcommand{\third}[1]{\cellcolor{thirdcolor}{#1}}

\algnewcommand{\Input}{\item[\textbf{Input:}]}

\title{Unified Loss-Aware Density Control\\
for 3D Gaussian Splatting}

\author{%
  Tae-Young Kim \quad Jiwoo Chung \quad Jae-Pil Heo$^{\dagger}$ \\
  Sungkyunkwan University \\
  \texttt{\{vanmeruso, jiwoo.jg, jaepilheo\}@gmail.com}
}
\newcommand{\jw}[1]
{
	\textcolor{red}{\bfseries{{JW:}}}
}

\begin{document}

\maketitle

\begingroup
\renewcommand{\thefootnote}{\ensuremath{\dagger}}
\footnotetext{Corresponding author}
\endgroup

\begin{abstract}

The efficiency and fidelity of 3D Gaussian Splatting critically depend on adaptive density control, which grows and suppresses Gaussian primitives throughout optimization. However, the standard pipeline still relies on heuristic view-space gradients, scale thresholds, and global opacity resets, rather than explicitly assessing the loss-reducing effect of each split, clone, or opacity update. This mismatch can allocate primitives to already-sufficient regions, miss under-reconstructed structures, and destabilize training through blanket opacity changes. In this work, we recast adaptive density control as a collection of local loss-reduction decisions. We derive efficient Taylor-based per-Gaussian surrogates that predict the objective change caused by three primitive-level operations: splitting, cloning, and opacity decay. This yields a unified loss-aware framework with three components. First, we extend second-order splitting to a normalized joint position-scale space, enabling children to adapt not only their locations but also their anisotropic extents. Second, we formulate duplicate cloning as a surrogate minimization problem, selecting only Gaussians whose duplicated contribution is predicted to improve reconstruction and assigning each child a principled opacity. Third, we replace global opacity reset with selective soft opacity decay that weakens harmful primitives while preserving useful ones. Across standard 3DGS benchmarks, our method produces more targeted primitive allocation and achieves improved rendering quality with comparable or smaller Gaussian budgets compared to baselines.

\end{abstract}

\section{Introduction}

% \begin{figure}[t!]
% \centering
% \includegraphics[width=1.0\textwidth]{figure/intro.pdf}
% \caption{\textbf{Intuition of our loss-aware adaptive density control~(ADC).}
% We visualize the effect of each density-control operation from the same intermediate state at the 3{,}000\(^{\text{th}}\) training iteration.
% (a) Overall comparison between conventional ADC and our method.
% (b) Splitting: we compare the position-only split of SteepGS~\cite{steep} with our position--scale split, visualizing Gaussians as ellipsoids. While the position-only split leaves gaps around the thin bicycle handle, our scale-aware split better adapts the offspring extent and forms the handle structure more faithfully.
% (c) For cloning, we compare standard 3DGS and ours by iteratively cloning candidate Gaussians and measuring the resulting loss change; green/red indicate candidates whose duplication decreases/increases the loss. Our cloning selects mostly loss-reducing duplicates, whereas heuristic cloning introduces noisier candidates.
% (d) For suppression, we compare the global opacity reset of standard 3DGS with our selective soft opacity reduction. Global opacity reset makes many Gaussians nearly transparent and requires additional re-optimization, making subsequent pruning less targeted, while our method selectively weakens Gaussians predicted to hinder optimization.
% }

% \label{fig:intro}
% \end{figure}

\begin{figure}[t!]
\centering
\includegraphics[width=1.0\textwidth]{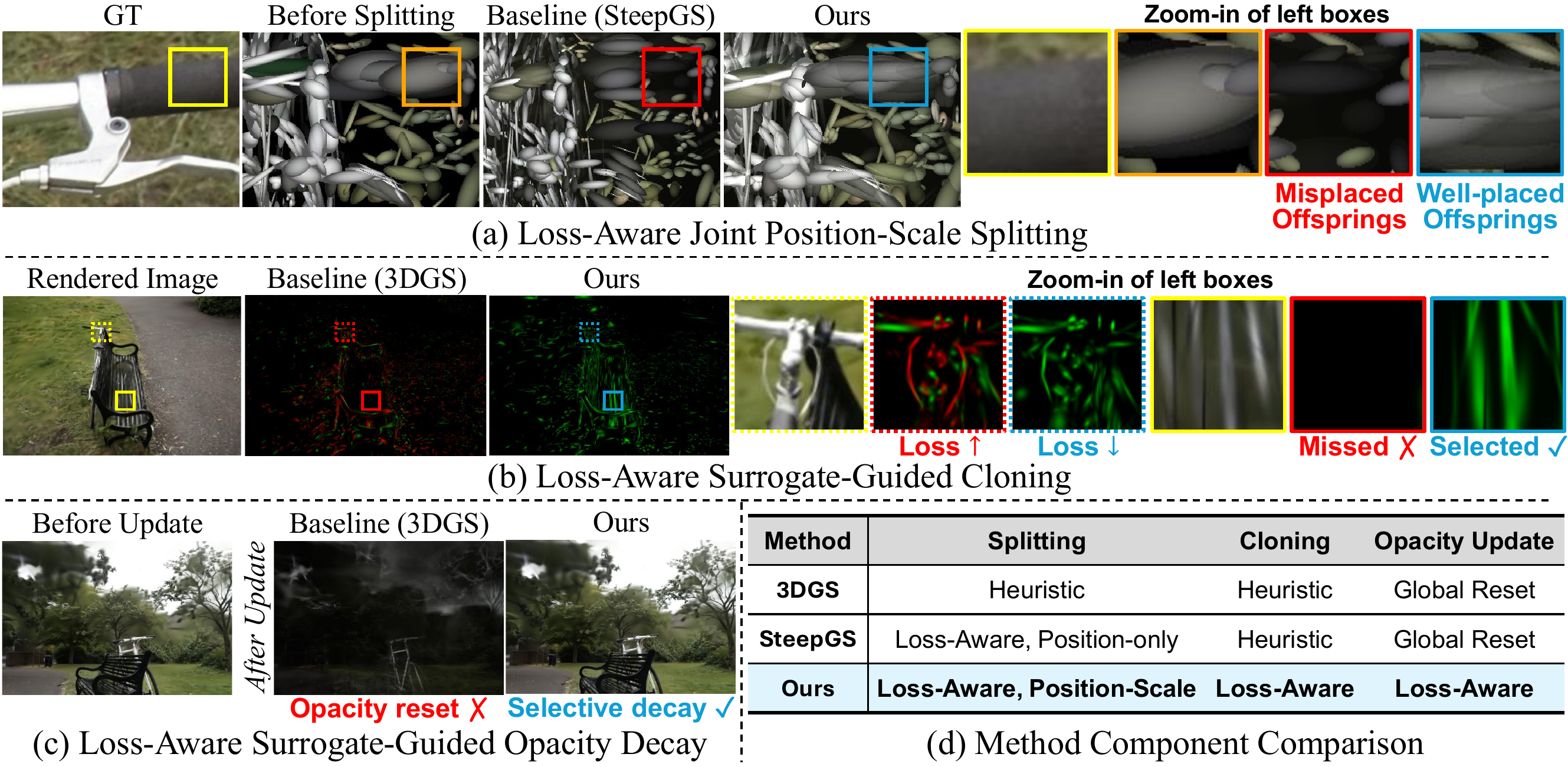}
\vspace{-12pt}
\caption{
\textbf{Intuition of our loss-aware adaptive density control~(ADC).}
% We visualize the effect of each density-control operation from the same intermediate state at the 3{,}000\(^{\text{th}}\) training iteration.
% (a) Splitting: we compare the position-only split of~\cite{steep} with our position-scale split, visualizing splitting Gaussians as ellipsoids. While the position-only split leaves gaps around the thin bicycle handle, our split better adapts the offspring extent and forms the handle structure more faithfully.
% (b) Cloning: we compare standard 3DGS and ours by \jw{individually} iteratively cloning candidate Gaussians and measuring the resulting loss change; \textcolor{ForestGreen}{green}/\textcolor{red}{red} indicate candidates whose duplication \textcolor{ForestGreen}{decreases}/\textcolor{red}{increases} the loss. Our cloning selects mostly loss-reducing duplicates, whereas heuristic cloning introduces noisier candidates.
% (c) Opacity decay: we compare the global opacity reset of standard 3DGS with our soft opacity decay. Global opacity reset makes all Gaussian primitives nearly transparent, leading to unstable training~\cite{revising}. In contrast, our method selectively weakens only Gaussians predicted to hinder optimization.
% (d) Overall comparison between conventional ADC and our method.
We visualize the effect of each density-control operation from the same intermediate state at the 3{,}000\(^{\text{th}}\) training iteration.
(a) Splitting: we compare the position-only split of~\cite{steep} with our position-scale split, visualizing splitting Gaussians as ellipsoids. While the position-only split produces misplaced offspring around the thin bicycle handle, our position-scale split places offspring more appropriately and forms the handle structure more faithfully.
(b) Cloning: we compare standard 3DGS and ours by individually cloning candidate Gaussians and measuring the resulting loss change; \textcolor{ForestGreen}{green}/\textcolor{red}{red} indicate candidates whose duplication \textcolor{ForestGreen}{decreases}/\textcolor{red}{increases} the loss. Our cloning selects mostly loss-reducing candidates and recovers missed structures, whereas heuristic cloning introduces loss-increasing candidates.
(c) Opacity decay: we compare the opacity reset of standard 3DGS with our selective opacity decay. Opacity reset makes all Gaussian primitives nearly transparent, leading to unstable training. In contrast, our method weakens only Gaussians predicted to hinder optimization.
(d) Component comparison: standard 3DGS relies on heuristics. SteepGS improves splitting with a position-only loss-aware rule but keeps cloning and opacity update heuristic. Our method applies loss-aware decisions to position-scale splitting, cloning, and opacity decay.
\vspace{-15pt}
}

\label{fig:intro}
\end{figure}
3D Gaussian Splatting (3DGS)~\cite{3dgs} has emerged as an effective scene representation for high-quality, real-time novel view synthesis. Subsequent studies have improved the reconstruction quality and rendering efficiency of 3DGS through advances in primitive representations~\cite{2dgs,betasplatting}, volumetric rendering~\cite{ever, vol3d}, anti-aliasing~\cite{mipsplatting, multigs}, and level-of-detail (LOD) strategies for large-scale scenes~\cite{hierarchicalgs}. 

Beyond these advances, reconstruction quality and representation complexity of 3DGS are also influenced by how Gaussian primitives are allocated and refined during training. Adaptive Density Control (ADC) is a key component of this process, progressively refining the scene representation through splitting, cloning, and pruning Gaussians. Despite its effectiveness, ADC still depends heavily on hand-crafted heuristic thresholds and positional gradients. These heuristic strategies can produce redundant Gaussian primitives even in regions where densification is unnecessary~\cite{revising, GDAGS, compact, steep}, which increases memory usage and slows rendering as they accumulate.

To address this problem, prior works have revisited ADC from various perspectives~\cite{absgs, dc4gs, GDAGS, mcmc, resgs, zhang2024fregs}. More recently, SteepGS~\cite{steep} mitigates redundant primitive generation by predicting whether splitting each Gaussian locally reduces the reconstruction loss. Despite these advances, two important aspects remain underexplored. In particular, existing loss-based splitting formulations mainly focus on position parameters, while the influence of other Gaussian parameters remains underexplored. Moreover, this loss-aware perspective has not yet been systematically extended to other density-control operations, such as cloning and opacity-based pruning, where conventional heuristic rules are still commonly used.

Building on these observations, we propose a loss-aware density control framework for efficient and faithful Gaussian representation. 
We first revisit splitting.
While moving offspring centers refines local geometry, thin or anisotropic structures often require adapting offspring spatial extents in addition to positions.
Thus, we extend splitting from a position-only formulation to a joint position-scale space, allowing each split to account for Gaussian extent. 
As shown in Fig.~\ref{fig:intro}~(a), position-only splitting can leave gaps around thin structures such as a bicycle handle, whereas our scale-aware splitting jointly adjusts offspring positions and scales to form a more faithful local representation.

We further extend the loss-aware principle to cloning.
Conventional cloning relies on view-space gradient and scale thresholds~\cite{3dgs}, without explicitly estimating whether a duplicate will reduce the photometric objective.
Our surrogate-guided cloning instead predicts the loss change induced by each duplicate and clones only primitives whose duplicate contributions are expected to reduce the loss.
As shown in Fig.~\ref{fig:intro}~(b), standard 3DGS selects both \textcolor{ForestGreen}{green} and \textcolor{red}{red} candidates in the zoom-in boxes.
This mixed selection reveals that its heuristic cloning rule is not directly aligned with the loss effect of duplication.
The baseline also misses important candidates around under-reconstructed structures such as the wooden frame, leaving these regions inadequately refined.
By contrast, our method selects mostly \textcolor{ForestGreen}{green} candidates and places clones in under-reconstructed regions, allowing the proposed surrogate to guide cloning toward primitives that are more likely to improve the objective.

A similar objective-agnostic issue appears in opacity-based pruning.
Standard 3DGS periodically applies a global opacity reset, reducing the opacity of all Gaussians regardless of whether their current contributions help or harm the objective.
To address this, our loss-aware opacity decay predicts which Gaussians currently increase the loss and weakens only those primitives, while preserving useful Gaussians.
As shown in Fig.~\ref{fig:intro}~(c), the global opacity reset makes all Gaussian primitives nearly transparent and introduces abrupt changes that can destabilize training~\cite{revising}, whereas our opacity decay avoids unnecessary global perturbations and yields a more stable training trajectory.

However, directly evaluating the loss change of each candidate operation is computationally expensive, as it requires applying the operation, re-rendering affected views, and recomputing the loss.
We therefore derive efficient Taylor-based per-Gaussian surrogates that approximate the objective change induced by splitting, cloning, and opacity decay~\cite{lecun1989optimal, wu2020firefly, molchanov2016pruning, molchanov2019importance}.
These surrogates provide a unified criterion for deciding whether each density-control operation is locally expected to reduce the loss.

Concretely, as shown in Fig.~\ref{fig:intro}~(d), our framework consists of three loss-aware components: position-scale splitting, surrogate-guided cloning, and surrogate-guided opacity decay.
The first adapts both offspring positions and spatial extents, the second selects loss-reducing duplicates and assigns adequate child opacity, and the third weakens Gaussians predicted to increase the objective.
Together, these components form a unified loss-aware ADC strategy that allocates primitives more effectively and improves rendering quality with fewer Gaussians on standard 3DGS benchmarks.

\section{Related works}
\subsection{Novel view synthesis}
Novel View Synthesis (NVS)~\cite{nvs} aims to render images of a 3D scene from previously unseen viewpoints using given multi-view observations. Neural Radiance Fields (NeRF)~\cite{nerf} have emerged as a major milestone in NVS by representing scenes as continuous volumetric fields. However, their reliance on dense sampling and implicit representations often leads to high rendering costs. In contrast, 3D Gaussian Splatting (3DGS)~\cite{3dgs} represents a scene explicitly as a set of anisotropic 3D Gaussians and combines this representation with differentiable, visibility-aware rasterization, enabling both efficient rendering and high-quality novel view synthesis.  

Recent studies have further improved 3DGS through alternative primitive representations~\cite{2dgs,betasplatting}, exact volumetric rendering formulations~\cite{vol3d,ever}, anti-aliasing techniques~\cite{mipsplatting,multigs}, and hierarchical level-of-detail (LOD) strategies~\cite{hierarchicalgs}. As a result of these advances, 3DGS has been actively studied across a wide range of tasks and application domains~\cite{inverserendering,gaussiangrouping,diffusion_based_editing,gsslam,seamless}.

% \subsection{Adaptive Density Control in 3DGS}
% Adaptive density control is a core component of 3DGS, determining which Gaussian primitives are split, cloned, or pruned during training. Because this process critically affects both reconstruction quality and primitive efficiency, a number of works have sought to make it more accurate. One line of work refines the signals used for densification by refining view-space positional gradients, introducing pixel-aware weighting, or exploiting directional consistency~\cite{absgs, pixelgs, dc4gs, GDAGS}. Others reinterpret adaptive density control through stochastic sampling frameworks, including MCMC-based state transitions and Metropolis-Hastings sampling~\cite{mcmc, mh}. A separate line of work instead modifies the densification procedure itself, for example through pixel-error-driven criteria or alternative residual-style splitting strategies~\cite{revising, resgs}. Most closely related to our work, SteepGS~\cite{steep} provides a loss-aware view of Gaussian splitting by deriving split directions from second-order information computed with respect to Gaussian positions.

% However, SteepGS applies this loss-aware treatment only to splitting and constructs the splitting criterion using only positional parameters. In contrast, our method extends this position-only splitting formulation to the joint position-scale space and further introduces loss-aware decision rules for cloning and opacity reduction, enabling a more principled treatment of the overall ADC process and leading to more effective density control.

\subsection{Adaptive Density Control in 3DGS}

Adaptive density control (ADC) determines how Gaussian primitives are cloned, split, and pruned during 3DGS training, and therefore strongly affects both reconstruction quality and primitive efficiency. Since the original 3DGS relies on heuristic gradient and scale thresholds, recent works have improved the split and clone operations in ADC by refining densification signals~\cite{absgs, pixelgs, dc4gs, GDAGS}, introducing stochastic sampling-based updates~\cite{mcmc, mh}, or revising the densification procedure itself~\cite{fang2024mini, revising, dongreframing, resgs}.

The most relevant work to ours is SteepGS~\cite{steep}, which makes Gaussian splitting loss-aware using second-order information. However, its formulation is limited to position-only splitting, while other ADC operations such as cloning and opacity reset also remain heuristic. We extend this loss-aware perspective to the broader ADC pipeline through joint position-scale splitting and local surrogate-based cloning and opacity reduction.

\section{Preliminary}
\subsection{3D Gaussian Splatting}
3D Gaussian Splatting (3DGS) represents a scene as a collection of \(n\) learnable anisotropic Gaussians,
\begin{equation}
\theta = \{\theta^{(i)}\}_{i=1}^{n},
\qquad
\theta^{(i)} = \bigl(p^{(i)}, \Sigma^{(i)}, o^{(i)}, c^{(i)}\bigr),
\end{equation}
where \(p^{(i)} \in \mathbb{R}^{3}\), \(\Sigma^{(i)} \in \mathbb{S}_{+}^{3\times3}\), \(o^{(i)}\), and \(c^{(i)}\) denote the center position, covariance, opacity, and appearance attributes of the \(i^\text{th}\) Gaussian, respectively. Following the standard 3DGS parameterization, each covariance matrix is expressed as \(\Sigma^{(i)} = R^{(i)} S^{(i)} S^{(i)\top} R^{(i)\top}\), where \(R^{(i)} \in SO(3)\) is a rotation matrix and \(S^{(i)} = \mathrm{diag}(s^{(i)})\) is a diagonal scaling matrix with \(s^{(i)} \in \mathbb{R}_{+}^{3}\). Under a camera projection \(\varphi : \mathbb{R}^{3} \rightarrow \mathbb{R}^{2}\), these Gaussians are projected on the image plane and composited in front-to-back depth order to render the color \(C(x)\) at pixel \(x \in \mathbb{R}^{2}\).
\begin{equation}
\alpha^{(i)}(x)
=
o^{(i)}
\exp\!\left(
-\frac{1}{2}
\bigl(x-\varphi(p^{(i)})\bigr)^{\top}
\left(\Sigma_{\varphi}^{(i)}\right)^{-1}
\bigl(x-\varphi(p^{(i)})\bigr)
\right).
\end{equation}
% \vspace{-5pt}
\begin{equation}
T^{(i)}(x)
=
\prod_{j=1}^{i-1}\bigl(1-\alpha^{(j)}(x)\bigr),
\qquad
C(x)
=
\sum_{i=1}^{n} c^{(i)}\,\alpha^{(i)}(x)T^{(i)}(x),
\label{alpha-blending}
\end{equation}
where \(T^{(i)}(x)\) is the accumulated transmittance before the \(i^\text{th}\) Gaussian, \(\alpha^{(i)}(x)\) is its image-space opacity contribution, and \(\Sigma_{\varphi}^{(i)}\) is its projected covariance on the image plane. Then, Gaussian parameters are optimized by minimizing the following photometric loss:
\begin{equation}
\mathcal{L}(\theta)
=
(1-\lambda_{\mathrm{D\text{-}SSIM}})\,\mathcal{L}_{1}(\theta)
+
\lambda_{\mathrm{D\text{-}SSIM}}\,\mathcal{L}_{\mathrm{D\text{-}SSIM}}(\theta),
\end{equation}
where \(\mathcal{L}_{1}\), \(\mathcal{L}_{\mathrm{D\text{-}SSIM}}\), and \(\lambda_{\mathrm{D\text{-}SSIM}}\) denote the pixel-wise L1 loss, SSIM loss~\cite{ssim}, and control hyperparameter between the two losses, respectively.

\subsection{Adaptive Density Control}
\label{sec:prelim_adc}

Along with standard gradient-based optimization, 3DGS~\cite{3dgs} periodically employs Adaptive Density Control (ADC) to dynamically refine the set of Gaussians during training.
ADC determines which Gaussians should be cloned, split, or pruned, and therefore has a direct effect on both reconstruction quality and primitive efficiency.
In standard 3DGS, ADC first identifies Gaussians with large average view-space positional gradients as densification candidates. Among these candidates, Gaussians with small scales are cloned to improve local coverage, while Gaussians with large scales are split into smaller offspring. In addition, Gaussians with low opacity are pruned, and opacity values are periodically reset as a heuristic stabilization mechanism.

However, these ADC rules do not explicitly assess whether a candidate operation would reduce the training objective. Candidate Gaussians are selected using a hand-crafted threshold on the view-space gradient magnitude, and the choice between cloning and splitting is determined using a scale threshold. As a result, Gaussians may be densified even when they could be refined solely through ordinary gradient-based updates, thereby increasing the number of Gaussians unnecessarily.

\subsection{SteepGS}
SteepGS~\cite{steep} introduces a loss-aware splitting strategy to address the limitations of heuristic density control in 3DGS. It observes that Gaussians can become stationary even in insufficiently reconstructed regions, limiting further loss reduction through gradient-based updates. SteepGS addresses this limitation by selectively splitting such Gaussians and displacing their offspring. To guide this process, it constructs a per-Gaussian splitting matrix that captures the second-order loss change induced by splitting. Using this matrix, it splits only Gaussians for which a local loss decrease is predicted. A detailed derivation of the splitting matrix is provided in Appendix~\ref{app:steepSplitting}.

However, its loss-aware splitting considers only Gaussian position, while cloning and opacity reset still rely on heuristic rules. We address these limitations by extending loss-aware splitting to the joint position-scale space and introducing surrogate-guided cloning and opacity decay.

\section{Methodology}
% 간략버전 수식
\begin{figure}[t]
    \centering
    \includegraphics[width=0.98\textwidth]{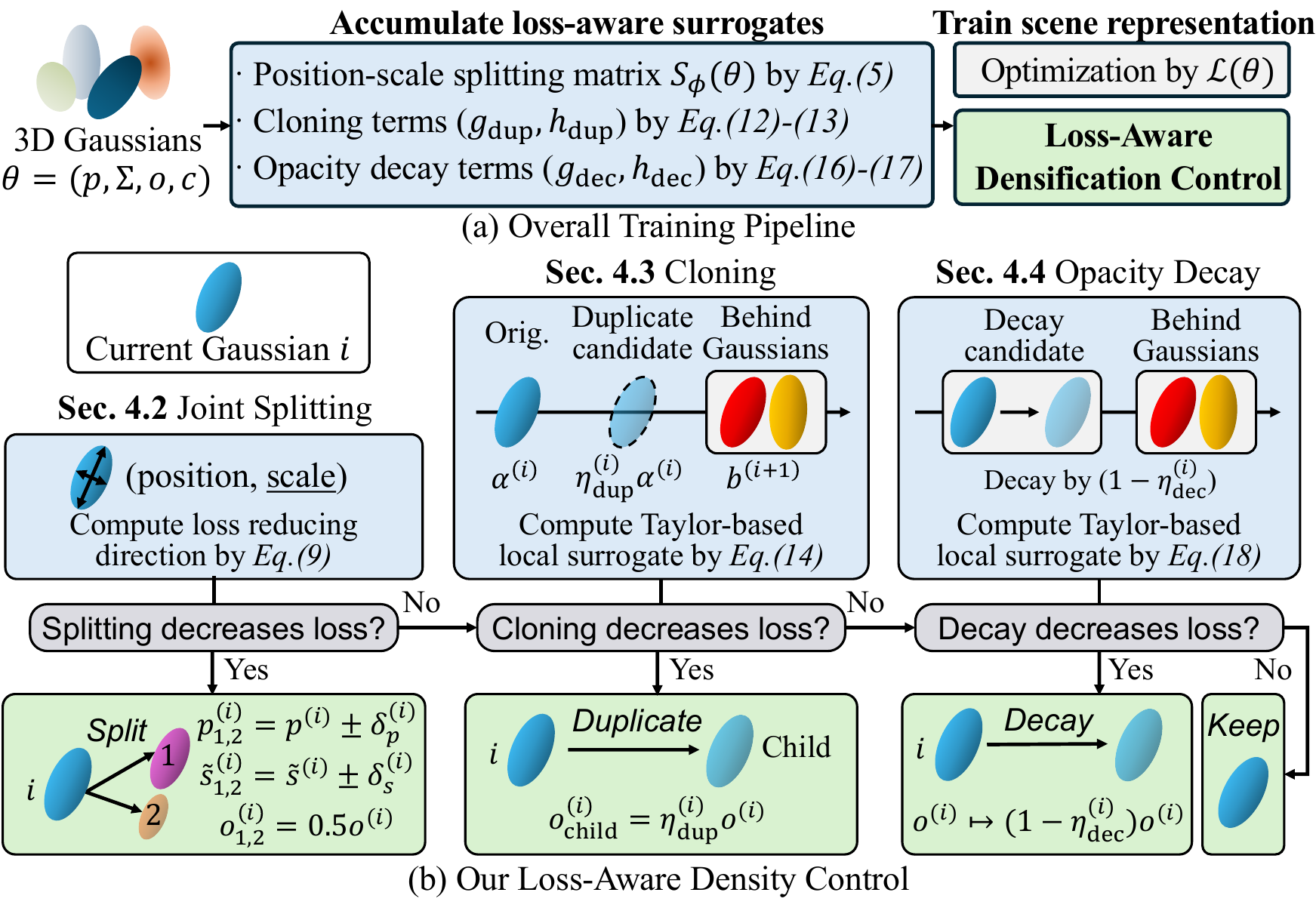}
    \vspace{-7pt}
 \caption{\textbf{Overview of our framework.}
(a) During training, we accumulate per-Gaussian surrogate statistics, including the joint splitting matrix and surrogate terms for cloning and opacity decay. At each densification interval, these surrogates determine whether each Gaussian is split, cloned, or assigned opacity decay.
(b) For each Gaussian, we sequentially evaluate splitting, cloning, and opacity decay using accumulated loss-aware surrogate statistics.  Joint splitting computes a loss-reducing direction in the joint position-scale space and initializes two children with corresponding position and log-scale perturbations. Cloning minimizes a local Taylor-based surrogate with respect to the duplicate opacity and appends a duplicate child when it is predicted to decrease the loss. 
Opacity decay similarly estimates a loss-reducing opacity decay strength and selectively decreases the opacity of Gaussians whose current contributions are predicted to be harmful.}
    \label{fig:main}
        \vspace{-5pt}
\end{figure}

\subsection{Overview}
\label{sec:method_overview}

Our goal is to replace the heuristic adaptive density control~(ADC) decisions with loss-aware criteria for splitting, cloning, and opacity decay.
As illustrated in Fig.~\ref{fig:main}, we accumulate per-Gaussian surrogate statistics during training and use them at each density-control interval to decide which operation should be applied to each Gaussian. Our framework consists of three components.
First, in Sec.~\ref{sec:method_split}, we extend loss-aware splitting from a position-only formulation to a normalized joint position-scale space, allowing offspring to adapt both their locations and spatial extents.
Second, in Sec.~\ref{sec:method_clone}, we introduce surrogate-guided cloning, which appends a duplicate child only when its contribution is predicted to decrease the loss and assigns its opacity from the same local surrogate.
Third, in Sec.~\ref{sec:method_opacity_decay}, we replace global opacity reset with surrogate-guided opacity decay, which weakens only Gaussians whose current contributions are predicted to increase the objective.
% The following sections describe these three operations: joint position-scale splitting in Sec.~\ref{sec:method_split}, surrogate-guided cloning in Sec.~\ref{sec:method_clone}, and surrogate-guided opacity decay in Sec.~\ref{sec:method_opacity_decay}.

% Our goal is to replace the heuristic Adaptive Density Control (ADC) rules of standard 3DGS with loss-aware decisions. An overview of our method is
% illustrated in Fig.~\ref{fig:main}.
% Our proposed method refines the Gaussian set through three loss-aware operations. First, we extend Hessian-guided splitting to a joint position-scale space, allowing split directions to reflect both positional sensitivity and Gaussian extent. Second, we introduce surrogate-guided cloning, which adds a child Gaussian only when duplication is locally favorable. Third, we replace the global opacity reset with selective soft opacity decay, suppressing Gaussians that are estimated to hinder optimization. 

% At each densification interval, these operations are applied in a prioritized sequence. We first identify Gaussians that should be split, then clone only the remaining ones that benefit from additional local capacity, and finally attenuate detrimental contributors. This converts ADC from a threshold-driven heuristic into a selective, loss-aware refinement process.

\subsection{Loss-Aware Joint Position-Scale Splitting}
\label{sec:method_split}
Our baseline SteepGS~\cite{steep} interprets Gaussian splitting from a loss-aware perspective and identifies directions that can reduce the reconstruction loss when each Gaussian is split. Specifically, SteepGS defines the split variable as the Gaussian position and therefore focuses only on offspring displacement in the position space. However, the rendering contribution of a Gaussian depends not only on its position but also on its scale, which determines its spatial extent. Therefore, effective splitting requires jointly adapting the positions and scales of the offspring to better capture local scene structure.

To formalize this idea, we build on the loss-aware splitting matrix of SteepGS~\cite{steep} and generalize its split variable from the position space to the joint position-scale space.
For each Gaussian \(i\), we define the splitting matrix
\(\mathcal{S}_{\phi}^{(i)}\), which measures the local effect of
splitting Gaussian \(i\) through the curvature of its image-space contribution
\(\alpha_i(x)\) with respect to a local split variable \(\phi^{(i)}\):
\begin{equation}
\phi^{(i)}
=
\begin{bmatrix}
p^{(i)} \\
\tilde s^{(i)}
\end{bmatrix},
\quad
\tilde s^{(i)}=\log s^{(i)},
\qquad
\mathcal{S}_{\phi}^{(i)}(\theta)
=
\mathbb{E}_{(\varphi,x)}
\left[
\frac{\partial \mathcal{L}_{\varphi,x}}
{\partial \alpha^{(i)}(x)}
\nabla_{\phi^{(i)}}^2
\alpha^{(i)}(x)
\right].
\label{eq:joint_split_matrix}
\end{equation}
where \(\tilde s^{(i)}\) denotes the log-scale vector associated with the
Gaussian scale \(s^{(i)}\), \(\mathcal{L}_{\varphi,x}\) denotes the per-pixel
photometric loss for camera \(\varphi\) and pixel \(x\), and the expectation is
taken over sampled training views and pixels. Since \(\phi^{(i)}\) contains both
position and log-scale variables, the resulting position-scale splitting matrix
\(\mathcal{S}_{\phi}^{(i)}(\theta)\) lies in \(\mathbb{R}^{6\times6}\). A detailed
derivation of the original SteepGS splitting matrix is provided in Appendix~\ref{app:steepSplitting}.

However, directly applying eigendecomposition to this raw matrix is inappropriate, since the position and log-scale blocks have different units and numerical scales, making their eigenvectors not directly comparable. We therefore evaluate split directions in a normalized joint coordinate system
as follows:
\begin{equation}
L^{(i)}
=
R^{(i)}\mathrm{diag}(s^{(i)}),
\qquad
J^{(i)}
=
\begin{bmatrix}
L^{(i)} & 0 \\
0 & \frac{1}{\sqrt{2}} \mathbf I
\end{bmatrix},
\qquad
\tilde{\mathcal{S}}_{\phi}^{(i)}
=
J^{(i)\top}
\bar{\mathcal{S}}_{\phi}^{(i)}
J^{(i)}.
\label{eq:joint_metric_transform}
\end{equation}
where \(J^{(i)}\) maps a normalized joint direction to a perturbation in the
position-log-scale space, \(L^{(i)}\) is the covariance factor of the
\(i^\text{th}\) Gaussian satisfying \(L^{(i)}L^{(i)\top}=\Sigma^{(i)}\), and
\(\mathbf I\) is the \(3\times3\) identity matrix.
This transformation enables split directions to be compared in a normalized
joint position-scale space. The detailed derivation of
Eq.~\ref{eq:joint_metric_transform} is provided in
Appendix~\ref{app:joint_pos_scale_matrix}. In this normalized space, we determine the split direction by computing the
minimum eigenpair of \(\tilde{\mathcal{S}}_{\phi}^{(i)}\):
\begin{equation}
(\lambda_{\mathrm{min}}^{(i)}, v_{\mathrm{min}}^{(i)})
=
\operatorname{eigmin}
\left(
\tilde{\mathcal{S}}_{\phi}^{(i)}
\right),
\label{eq:eigmin_split}
\end{equation}
where \(\lambda_{\mathrm{min}}^{(i)}\) is the smallest eigenvalue and
\(v_{\min}^{(i)}\) is the corresponding normalized joint direction. Since
\(v_{\min}^{(i)}\) gives the direction of smallest predicted second-order loss
change, a negative \(\lambda_{\min}^{(i)}\) indicates a local loss-decreasing
split direction. This is especially useful for nearly stationary Gaussians,
where gradient-based updates make little progress and the negative-curvature
direction moves the offspring away from the stationary region~\cite{steep}.
We therefore split Gaussian \(i\) only when it is nearly stationary, has
unresolved view-space variation, and exhibits sufficiently negative curvature:
\begin{equation}
\mathcal I_{\mathrm{split}}
=
\Bigl\{\, i \,\Bigm|\,
\left(\|\bar g_{\mathrm{pos}}^{(i)}\|_2 \le \tau_{\mathrm{pos}}\right)
\land
\left(\|\bar g_{\mathrm{scale}}^{(i)}\|_2 \le \tau_{\mathrm{scale}}\right)
\land
\left(\lambda_{\mathrm{min}}^{(i)} < \tau_{\mathrm{split}}\right)
\Bigr\}.
\label{eq:split_candidate}
\end{equation}
where \(\mathcal I_{\mathrm{split}}\) denotes the set of Gaussians selected for
splitting, and \(\bar g_{\mathrm{pos}}^{(i)}\) and
\(\bar g_{\mathrm{scale}}^{(i)}\) are the position and log-scale
gradients of Gaussian \(i\), respectively. The thresholds
\(\tau_{\mathrm{pos}}\) and \(\tau_{\mathrm{scale}}\) are used to test whether
the Gaussian is nearly stationary with respect to position and scale updates.
The negative threshold \(\tau_{\mathrm{split}}\) prevents splits caused by noisy
curvature arising from the approximation process. Once a Gaussian satisfies these criteria, its offspring are initialized along
the joint split direction:
\begin{equation}
p_{1,2}^{(i)}
=
p^{(i)}
\pm
\delta_p^{(i)} ,
\quad
\tilde s_{1,2}^{(i)}
=
\tilde s^{(i)}
\pm
\delta_{\tilde s}^{(i)},
\quad
o_{1,2}^{(i)}
=
\frac{1}{2}
o^{(i)},
\quad
\begin{bmatrix}
\delta_p^{(i)} \\
\delta_{\tilde s}^{(i)}
\end{bmatrix}
=
\delta \phi^{(i)}
=
J^{(i)}v_{\mathrm{min}}^{(i)},
\label{eq:split_update_joint}
\end{equation}

where \(\delta_p^{(i)}\) and \(\delta_{\tilde s}^{(i)}\) are the position and log-scale displacements. The two children inherit the parent rotation and appearance. By defining the split direction in the joint position-scale space, our splitting rule estimates not only loss-aware position displacements \(\delta p^{(i)}\), but also loss-aware scale refinements \(\delta \tilde s^{(i)}\), enabling the offspring to better adapt to the local reconstruction geometry.

\subsection{Loss-Aware Surrogate-Guided Cloning}
\label{sec:method_clone}
We next introduce local surrogate-based rules for cloning and opacity decay. Although the two operations are different, they both modify the opacity contribution of a Gaussian. Specifically, cloning adds a new contribution by appending an exact duplicate at the same location, whereas opacity decay removes part of the current contribution. This shared structure allows both operations to be modeled with a quadratic approximation around zero operation strength. For Gaussian \(i\), we define the cloning operation \(\mathcal C(\theta;\eta^{(i)}_{\mathrm{dup}})\)  following the standard 3DGS, except
that the duplicate child's opacity is explicitly parameterized by a relative
opacity variable \(\eta^{(i)}_{\mathrm{dup}}\):
\begin{equation}
\mathcal C(\theta;\eta^{(i)}_{\mathrm{dup}})
=
\bigl(
p^{(i)}, \Sigma^{(i)}, \eta^{(i)}_{\mathrm{dup}}\,o^{(i)}, c^{(i)}
\bigr),
\qquad
\eta^{(i)}_{\mathrm{dup}} \in [0,1],
\label{eq:clone_action}
\end{equation}
where \(\mathcal C(\theta;\eta^{(i)}_{\mathrm{dup}})\) denotes the duplicate child
candidate with relative opacity \(\eta^{(i)}_{\mathrm{dup}}\). When \(\eta^{(i)}_{\mathrm{dup}}=1\), it
coincides with standard 3DGS cloning. To derive a local surrogate, we analyze
its image-space effect. Following Eq.~\ref{alpha-blending}, we define the image-space effect of Gaussian \(i\) at pixel \(x\) as \(\Delta^{(i)}(x)\), and the
per-unit duplicate contribution as \(\Delta_{\mathrm dup}^{(i)}(x)\):
\begin{equation}
\Delta^{(i)}(x)
=
T^{(i)}(x)\,\alpha^{(i)}(x)\,
\bigl(
c^{(i)}-b^{(i+1)}(x)
\bigr),
\qquad
\Delta_{\mathrm dup}^{(i)}(x)
=
\bigl(
1-\alpha^{(i)}(x)
\bigr)
\Delta^{(i)}(x),
\label{eq:clone_delta_terms}
\end{equation}
where \(b^{(i+1)}(x)\) denotes the background-inclusive color
composited from all Gaussians behind Gaussian \(i\).
Specifically, \(\Delta^{(i)}(x)\) measures the image-space effect induced by
Gaussian \(i\), and \(\Delta_{\mathrm{dup}}^{(i)}(x)\) denotes the image-space change caused by
adding a duplicate of Gaussian \(i\) with unit opacity weight. Thus, assigning opacity
\(\eta^{(i)}_{\mathrm{dup}}\) to the child induces the perturbation
\(\eta^{(i)}_{\mathrm{dup}}\Delta_{\mathrm{dup}}^{(i)}(x)\).
Using this perturbation, we define a Taylor-based local surrogate for cloning:
\begin{gather}
\mathcal{L}\!\left(
\theta
\cup
\left\{
\mathcal C(\theta;\eta^{(i)}_{\mathrm{dup}})
\right\}
\right)
=
\mathcal{L}(\theta)
+
{\eta^{(i)}_{\mathrm{dup}}}\,g^{(i)}_{\mathrm{dup}}
+
\frac{1}{2}({\eta^{(i)}_{\mathrm{dup}}})^2 h^{(i)}_{\mathrm{dup}}
+
O((\eta^{(i)}_{\mathrm{dup}})^3)
\label{eq:clone_quadratic}
\\
\begin{aligned}
g^{(i)}_{\mathrm{dup}}
&=
\sum_{x,c}
\frac{\partial \mathcal{L}}{\partial I_c(x)}
\Delta_{\mathrm{dup},c}^{(i)}(x),
\qquad
h^{(i)}_{\mathrm{dup}}
=
\sum_{x,c}
\omega_c(x)
\left(
\Delta_{\mathrm{dup},c}^{(i)}(x)
\right)^2,
\end{aligned}
\label{eq:clone_gh}
\end{gather}
where the summation is over all image pixels \(x\) and color channels \(c\), with nonzero terms only for pixels affected by Gaussian \(i\). \(I_c(x)\) denotes the rendered color value at pixel \(x\) in channel \(c\), \(\Delta_{\mathrm{dup},c}^{(i)}(x)\) is the corresponding channel of \(\Delta_{\mathrm{dup}}^{(i)}(x)\), \(\omega_c(x)\) denotes the proxy curvature weight described in Appendix~\ref{sec:residual_curvature}, and \(O((\eta^{(i)}_{\mathrm{dup}})^3)\) denotes the omitted higher-order remainder.

This surrogate estimates the loss variation induced by assigning a small relative opacity to the duplicate child. Based on this local approximation, we determine whether a duplicate can reduce the current loss by minimizing the quadratic surrogate with respect to the clone strength. This gives
the predicted strength 
\(\eta^{(i)}_{\mathrm{dup}}=-g^{(i)}_{\mathrm{dup}}/h^{(i)}_{\mathrm{dup}}\).
We append the duplicate child with strength \(\eta^{\mathrm{dup}}\) only when the local
surrogate predicts a valid loss decrease within the small-perturbation regime:
\begin{equation}
\begin{aligned}
\mathcal I_{\mathrm{clone}}
=
\Bigl\{\, i \,\Bigm|\,
&\left(i \notin \mathcal I_{\mathrm{split}}\right)
\land
\left(g^{(i)}_{\mathrm{dup}} < 0\right)
\land
\left(\tau_{\mathrm{clone}} < \eta^{(i)}_{\mathrm{dup}} \le 10^{-2}\right)
\Bigr\}, \\
\theta
&\leftarrow
\theta
\cup
\left\{
\bigl(
p^{(i)}, \Sigma^{(i)}, \eta^{(i)}_{\mathrm{dup}}\,o^{(i)}, c^{(i)}
\bigr)
\,\middle|\,
i \in \mathcal I_{\mathrm{clone}}
\right\}.
\end{aligned}
\label{eq:clone_set_update}
\end{equation}
where \(\mathcal I_{\mathrm{clone}}\) denotes the set of Gaussians selected for
duplicate cloning. The threshold \(\tau_{\mathrm{clone}}\) filters out
negligible clone strengths, while the upper bound restricts the duplicate
operation to a small perturbation regime in which the Taylor-based surrogate
remains reliable. This selective cloning expands only Gaussians whose duplicate contribution is predicted to reduce the loss, helping avoid redundant Gaussian generation.

%%%%%%%%%%%%%%%%%%%%%%%%%%%%%%%%%%%%%%%%%%%%%
\subsection{Loss-Aware Surrogate-Guided Opacity Decay}
\label{sec:method_opacity_decay}

As the counterpart to cloning, opacity decay decreases part of the current
opacity contribution of a Gaussian. Instead of applying a global opacity reset,
we selectively reduce the opacity of Gaussians whose current contributions are
predicted to hinder optimization. For Gaussian \(i\), we define the opacity decay operation
\(\mathcal P(\theta;\eta^{(i)}_{\mathrm{dec}})\) by reducing only its opacity while
preserving the other parameters:
\begin{equation}
\mathcal P(\theta;\eta^{(i)}_{\mathrm{dec}})
=
\theta
\bigl[
o^{(i)}
\mapsto
(1-\eta^{(i)}_{\mathrm{dec}})o^{(i)}
\bigr],
\qquad
\eta^{(i)}_{\mathrm{dec}}\in[0,1].
\label{eq:prune_action}
\end{equation}
where \(\eta^{(i)}_{\mathrm{dec}}\) denotes the fraction by which the opacity is reduced. To derive a local surrogate, we use the image-space effect
\(\Delta^{(i)}(x)\) defined in Eq.~\ref{eq:clone_delta_terms}. Unlike cloning, opacity decay does not introduce a new Gaussian or an additional image-space term, but simply reduces the existing contribution of the selected Gaussian.
Therefore, applying opacity decay with strength \(\eta^{(i)}_{\mathrm{dec}}\) to
Gaussian \(i\) induces the image-space perturbation
\(-\eta^{(i)}_{\mathrm{dec}}\Delta^{(i)}(x)\) at pixel \(x\).
Using this perturbation, we define a
Taylor-based local surrogate for opacity decay:
\begin{gather}
\mathcal{L}\!\left(\mathcal P(\theta;{\eta^{(i)}_{\mathrm{dec}}})\right)
=
\mathcal{L}(\theta)
-
\eta^{(i)}_{\mathrm{dec}}\,g^{(i)}_{\mathrm{dec}}
+
\frac{1}{2}{\eta^{(i)}_{\mathrm{dec}}}^2 h^{(i)}_{\mathrm{dec}}
+
O((\eta^{(i)}_{\mathrm{dec}})^3)
\label{eq:prune_quadratic}
\\
\begin{aligned}
g^{(i)}_{\mathrm{dec}}
&=
\sum_{x,c}
\frac{\partial \mathcal{L}}{\partial I_c(x)}
\Delta_{c}^{(i)}(x),
\qquad
h^{(i)}_{\mathrm{dec}}
=
\sum_{x,c}
\omega_c(x)
\left(
\Delta_{c}^{(i)}(x)
\right)^2,
\end{aligned}
\label{eq:prune_gh}
\end{gather}
where \(\Delta_{c}^{(i)}(x)\) denotes the \(c^\text{th}\) channel of \(\Delta^{(i)}(x)\). To determine whether reducing the opacity of Gaussian primitive \(i\) can locally reduce
the current loss, we follow the same local minimization as in cloning. This gives the predicted decay strength
\(\eta^{(i)}_{\mathrm{dec}}=g^{(i)}_{\mathrm{dec}}/h^{(i)}_{\mathrm{dec}}\).
We then apply opacity decay only when this strength is positive and remains
within a small perturbation range:
\begin{equation}
\begin{aligned}
\mathcal I_{\mathrm{decay}}
=
\Bigl\{\, i \,\Bigm|\,
&\left(i \notin \mathcal I_{\mathrm{split}}\right)
\land
\left(i \notin \mathcal I_{\mathrm{clone}}\right)
\land
\left(g^{(i)}_{\mathrm{dec}} > 0\right)
\land
\left(\tau_{\mathrm{decay}} < \eta^{(i)}_{\mathrm{dec}} \le 10^{-2}\right)
\Bigr\}, \\
\theta
&\leftarrow
\theta
\bigl[
o^{(i)}
\mapsto
(1-\eta^{(i)}_{\mathrm{dec}})o^{(i)}
\;\bigm|\;
i \in \mathcal I_{\mathrm{decay}}
\bigr].
\end{aligned}
\label{eq:soft_prune_set_update}
\end{equation}
where \(\mathcal I_{\mathrm{decay}}\) denotes the set of Gaussians selected for
opacity decay. The threshold \(\tau_{\mathrm{decay}}\) filters out
negligible opacity-decay strengths, while the upper bound restricts the
opacity update to a sufficiently local change where the Taylor-based surrogate
is expected to remain accurate. This selective formulation reduces opacity only for Gaussians whose current contributions are predicted to increase the loss. In contrast to the global opacity reset used in 3DGS, it provides a more targeted update and avoids
abrupt discontinuities in the training trajectory.

\section{Experiments}

\subsection{Implementation Details}
\label{implementation_detail}

\noindent\textbf{Benchmarks.}
We evaluate our method on three widely used real-world benchmarks for 3D Gaussian Splatting, namely Mip-NeRF360~\cite{mipnerf360}, Tanks\&Temples~\cite{tankandtemples}, and Deep Blending~\cite{Deepblending}.
Following the standard 3DGS evaluation protocol, we use all scenes from Mip-NeRF360, the Train and Truck scenes from Tanks\&Temples, and the DrJohnson and Playroom scenes from Deep Blending.
We use the image resolution settings commonly adopted in prior 3DGS works. Specifically, outdoor scenes in Mip-NeRF360 are downsampled by a factor of 4, indoor scenes by a factor of 2, and the remaining datasets are evaluated at their original resolution.
% We evaluate our method on three widely used real-world benchmarks for 3D Gaussian Splatting: Mip-NeRF360~\cite{mipnerf360}, Tanks\&Temples~\cite{tankandtemples}, and Deep Blending~\cite{Deepblending}. Following the standard 3DGS evaluation protocol, we use all scenes from Mip-NeRF360, the Train and Truck scenes from Tanks\&Temples, and the DrJohnson and Playroom scenes from Deep Blending.
% For all experiments, we follow the standard image resolution setting used in prior 3DGS works. Specifically, outdoor scenes in Mip-NeRF360 are downsampled by a factor of 4, indoor scenes by a factor of 2, and the remaining datasets are evaluated at their original image resolution.

\noindent\textbf{Training configuration.}
Our training setup follows the original 3DGS configuration~\cite{3dgs} and uses the same hyperparameters across all scenes and benchmarks, with detailed settings provided in Appendix~\ref{app:hyperparameters}.
All experiments are conducted on a single NVIDIA A6000 GPU.
For evaluation, we report PSNR, SSIM~\cite{ssim}, LPIPS~\cite{Lpips}, and the average number of Gaussians.
To ensure a fair comparison, we reproduce all baselines using their official implementations under the same software environment and evaluation protocol.
Each experiment is repeated five times, and we report the mean and
standard deviation across runs, with the standard deviation rounded
to two decimal places.
% Our experimental setup follows the training configuration of the original 3DGS~\cite{3dgs}. Our hyperparameters are kept identical across all experiments, and the detailed settings of experiments are provided in the Appendix~\ref{app:hyperparameters}. All experiments are conducted on a single NVIDIA A6000 GPU. For evaluation, we report PSNR, SSIM, LPIPS~\cite{Lpips} and the average number of Gaussians for each benchmark. To ensure a fair comparison across methods, we reproduced all baselines using
% the official publicly released code under the same software environment and evaluation protocol. Each experiment was repeated five times,
% and we report the mean and standard deviation rounded to two decimal places to account for random variation across runs.

\begin{table}[t]
\caption{
% Quantitative comparison on standard 3DGS benchmarks. Num denotes the average number of Gaussians over all scenes in each dataset and is reported in millions. 
% $^{*}$ denotes results obtained with 3DGS + residual splitting following the setting described in the corresponding paper~\cite{resgs}, to isolate the effect of the proposed densification method.
Quantitative comparison on standard 3DGS benchmarks. Num denotes average Gaussians (M); $^{*}$ follows the 3DGS+residual-splitting setting of~\cite{resgs} to isolate the effect of the proposed method.
}
\vspace{5pt}
\centering
\scriptsize
\setlength{\tabcolsep}{3pt}
\renewcommand{\arraystretch}{1.1}
\resizebox{\columnwidth}{!}{%
\begin{tabular}{c|cccc|cccc|cccc}
\hline
Dataset
& \multicolumn{4}{c|}{Mip-NeRF360}
& \multicolumn{4}{c|}{Tanks\&Temples}
& \multicolumn{4}{c}{Deep Blending} \\
Method
& PSNR$\uparrow$ & SSIM$\uparrow$ & LPIPS$\downarrow$ & Num
& PSNR$\uparrow$ & SSIM$\uparrow$ & LPIPS$\downarrow$ & Num
& PSNR$\uparrow$ & SSIM$\uparrow$ & LPIPS$\downarrow$ & Num \\
\hline
3DGS~\cite{3dgs} 
& $27.35$ & $0.814$ & $0.217$ & $3.4$ 
& $23.39$ & $0.841$ & $0.183$ & $1.8$ 
& $29.52$ & $0.903$ & $0.242$ & $3.0$ \\
\hline

AbsGS~\cite{absgs}
& $27.47_{\pm0.04}$ & $0.819_{\pm0.00}$ & $\third{0.191}_{\pm0.00}$ & $3.1$ 
& $23.67_{\pm0.06}$ & $0.852_{\pm0.00}$ & $\third{0.163}_{\pm0.00}$ & $1.3$ 
& $\third{29.62}_{\pm0.04}$ & $0.901_{\pm0.00}$ & $\third{0.236}_{\pm0.00}$ & $1.9$ \\

PixelGS~\cite{pixelgs}
& $27.58_{\pm0.01}$ & $\second{0.824}_{\pm0.00}$ & $\second{0.190}_{\pm0.00}$ & $5.5$ 
& $23.78_{\pm0.04}$ & $\best{0.857}_{\pm0.00}$ & $\best{0.150}_{\pm0.00}$ & $4.5$ 
& $28.93_{\pm0.03}$ & $0.897_{\pm0.00}$ & $0.248_{\pm0.00}$ & $4.6$ \\

DC4GS~\cite{dc4gs}  
& $\third{27.59}_{\pm0.02}$ & $\best{0.826}_{\pm0.00}$ & $\best{0.189}_{\pm0.00}$ & $2.6$ 
& $\third{23.94}_{\pm0.04}$ & $\second{0.857}_{\pm0.00}$ & $\second{0.161}_{\pm0.00}$ & $1.1$ 
& $\second{29.65}_{\pm0.03}$ & $\best{0.906}_{\pm0.00}$ & $\best{0.235}_{\pm0.00}$ & $1.5$ \\

GDAGS~\cite{GDAGS}  
& $27.50_{\pm0.01}$ & $\third{0.822}_{\pm0.00}$ & $0.192_{\pm0.00}$ & $3.0$ 
& $23.66_{\pm0.06}$ & $0.852_{\pm0.00}$ & $0.167_{\pm0.00}$ & $1.4$ 
& $29.61_{\pm0.05}$ & $0.904_{\pm0.00}$ & $\second{0.236}_{\pm0.00}$ & $2.0$ \\

ResGS$^{*}$~\cite{resgs}  
& $\second{27.65}_{\pm0.02}$ & $0.817_{\pm0.00}$ & $0.218_{\pm0.00}$ & $1.9$ 
& $\second{24.02}_{\pm0.03}$ & $0.851_{\pm0.00}$ & $0.178_{\pm0.00}$ & $1.0$ 
& $29.54_{\pm0.17}$ & $0.903_{\pm0.00}$ & $0.239_{\pm0.00}$ & $1.2$ \\

SteepGS~\cite{steep}
& $27.05_{\pm0.02}$ & $0.795_{\pm0.00}$ & $0.248_{\pm0.00}$ & $2.1$ 
& $23.48_{\pm0.04}$ & $0.839_{\pm0.00}$ & $0.193_{\pm0.00}$ & $1.3$ 
& $29.59_{\pm0.05}$ & $\third{0.904}_{\pm0.00}$ & $0.251_{\pm0.00}$ & $1.6$ \\
\hline
Ours   
& $\best{27.72}_{\pm0.02}$ & $0.818_{\pm0.00}$ & $0.219_{\pm0.00}$ & $1.9$ 
& $\best{24.20}_{\pm0.02}$ & $\third{0.853}_{\pm0.00}$ & $0.174_{\pm0.00}$ & $1.0$  
& $\best{29.79}_{\pm0.09}$ & $\second{0.906}_{\pm0.00}$ & $0.245_{\pm0.00}$ & $1.3$ \\
\hline
\end{tabular}
}
\label{tab:main_results}
\vspace{0pt}
\end{table}

\begin{figure}[t]
    \centering
    \includegraphics[width=\linewidth]{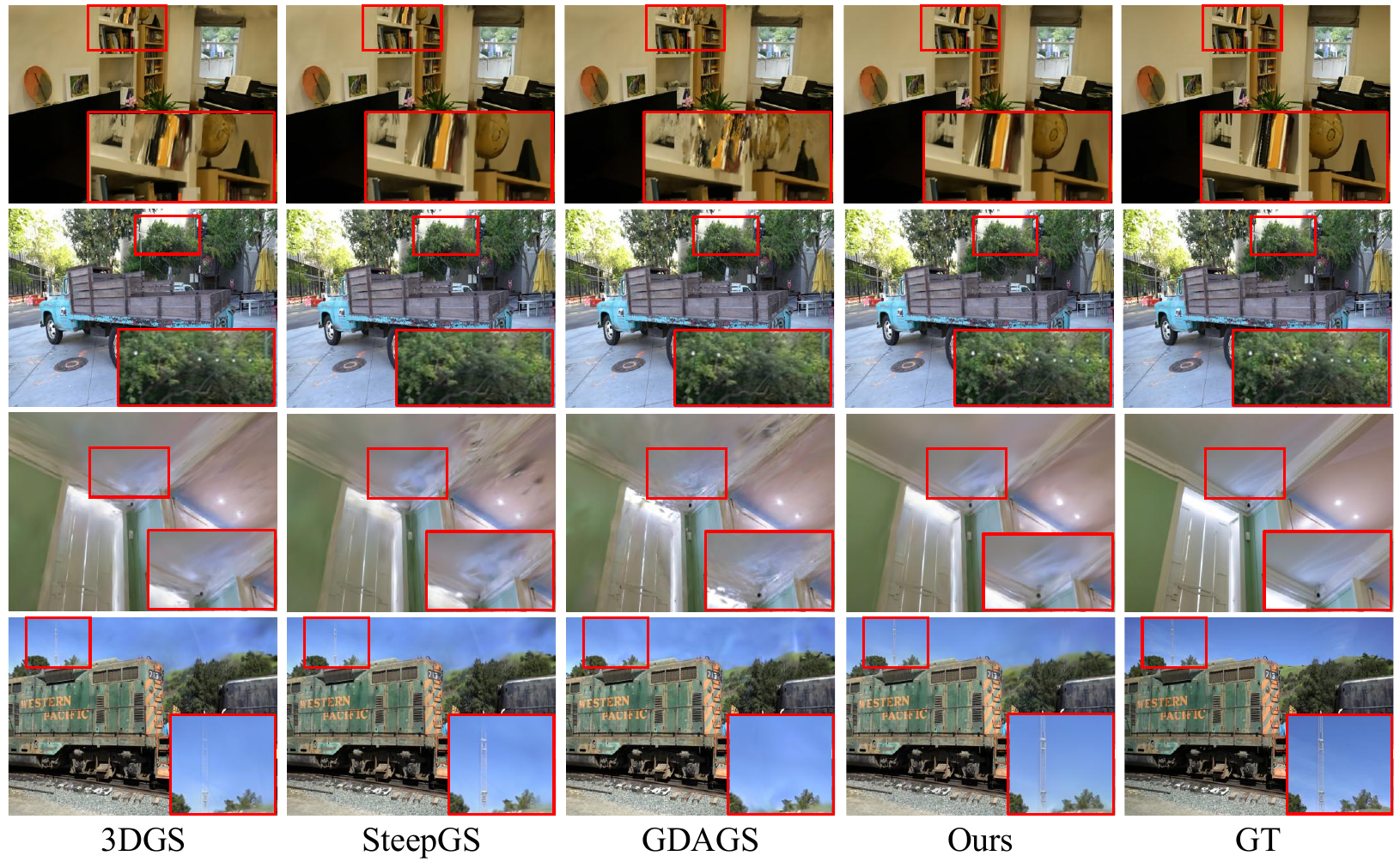}
    \vspace{-10pt}
    \caption{
    Qualitative comparison on standard 3DGS benchmarks. \textcolor{red}{Red} boxes highlight zoom-in regions where our method better preserves fine details and object boundaries.
    % \textbf{Qualitative results.} Compared with prior methods, our method achieves comparable or better reconstruction quality while using fewer Gaussians. The highlighted regions (red boxes) show that our method better preserves fine details and object boundaries.
    }
    \label{fig:qualitative}
    \vspace{-5pt}
\end{figure}

\subsection{Experimental Results}
% \subsection{Quantitative comparison}
% \jw{%기존 방법이랑 compare 했을때 -> densify가 필요한 점이랑 gradient update만으로 해결할 수 있는 점들을 구별하지 않고, redudnant하게 gaussian을 생성함, 반면 우리의 방법은 loss-aware하게 실제 densify가 필요하다고 생각되는 점에 대해서만 진행했기 떄문에, gaussian 개수를 compact하게 줄이고 좋은 성능을 얻을 수 있었음.
% % SteepGS랑 비교했을때도, 더 principle하게 해서 더 적거나 비슷한 개수로도 뛰어난 성능 향상을 낼 수 있었다.. 정도..?
% % mcmc는 흠.. 아직 실험이 덜끝나서 어떻게 말하쥐..

% }
% Tab.~\ref{tab:main_results} reports quantitative comparisons on standard 3DGS benchmarks.
% Our method achieves the best PSNR across all datasets while using substantially fewer Gaussians than standard 3DGS, showing a favorable quality-compactness trade-off.
% This suggests that heuristic ADC can allocate redundant primitives to regions that do not necessarily require densification, whereas our loss-aware surrogates select split, clone, and opacity-decay operations only when they are predicted to reduce the objective.
% Although DC4GS obtains marginally better SSIM and LPIPS in some cases, our method uses fewer Gaussians and achieves higher PSNR across all datasets, indicating a stronger reconstruction--compactness trade-off.
% Compared with SteepGS~\cite{steep}, the closest loss-aware density-control baseline, our method consistently improves reconstruction quality with a comparable or smaller Gaussian budget.
% These results demonstrate the benefit of extending loss-aware densification control beyond heuristics to the loss-aware selective ADC pipeline.

\noindent\textbf{Quantitative results.}
In Tab.~\ref{tab:main_results}, our method achieves the best PSNR across all datasets while using substantially fewer Gaussians than standard 3DGS.
Heuristic ADC can introduce redundant primitives in regions that do not necessarily require densification, whereas our loss-aware surrogates apply splitting, cloning, and opacity decay only when they are predicted to reduce the objective.
Although DC4GS~\cite{dc4gs} achieves marginally better SSIM and LPIPS in some cases, our method uses fewer Gaussians and obtains higher PSNR across all datasets, indicating a stronger reconstruction-compactness trade-off.
Compared with SteepGS~\cite{steep}, the closest loss-aware density-control baseline, our method consistently improves reconstruction quality with a smaller number of Gaussians.
These results validate the effectiveness of our loss-aware density control.

\noindent\textbf{Qualitative results.}
In Fig.~\ref{fig:qualitative}, our method preserves fine structures and object boundaries more faithfully than the baselines, even with a smaller Gaussian budget.
Our loss-aware density control allocates primitives to visually important regions rather than simply increasing the number of Gaussians.
For example, in the second row, the red highlighted region around the tree and white lamps shows that our method reconstructs all white lamps clearly.
However, the baselines tend to miss or blur some of the white lamps.
Similar improvements are observed in the red boxed regions of the last row, where our method successfully reconstructs the thin spire above the train, whereas the baselines produce blurry structures and partially fail to recover the detail.
These results show that our method efficiently allocates Gaussian primitives to regions that require additional capacity.
% , enabling more faithful local reconstruction under a compact Gaussian budget.

\subsection{Analysis}
% \subsection{Ablation Study}

\begin{table}[t]
\caption{Ablation study of the proposed components on Mip-NeRF360~\cite{mipnerf360}.}
\vspace{5pt}
\label{tab:ablation}
\centering
\setlength{\tabcolsep}{4pt}
\renewcommand{\arraystretch}{1.15}
\resizebox{\columnwidth}{!}{%
\begin{tabular}{c|cccc|c|cccc}
\hline
Method 
& PSNR$\uparrow$ & SSIM$\uparrow$ & LPIPS$\downarrow$ & Num
& Method
& PSNR$\uparrow$ & SSIM$\uparrow$ & LPIPS$\downarrow$ & Num \\
\hline
(a): Baseline~\cite{steep}
& $27.05$ & $0.795$ & $0.248$ & $2.1$
&
(c): (b) + Sec~\ref{sec:method_clone}
& $27.37$ & $0.813$ & $0.229$ & $\mathbf{1.6}$ \\

(b): (a) + Sec~\ref{sec:method_split}
& $27.10$ & $0.810$ & $0.230$ & $2.0$
&
(d): (c) + Sec~\ref{sec:method_opacity_decay}
& $\mathbf{27.72}$ & $\mathbf{0.818}$ & $\mathbf{0.219}$ & $1.9$ \\
\hline
\end{tabular}
}
\end{table}

\begin{table}[t]
\caption{
Direct comparison of position-only and joint position-scale splitting.
}
\label{tab:split_comparison}
\vspace{5pt}
\centering
\setlength{\tabcolsep}{3pt}
\renewcommand{\arraystretch}{1.1}
\resizebox{\columnwidth}{!}{%
\begin{tabular}{c|cccc|cccc|cccc}
\hline
Dataset
& \multicolumn{4}{c|}{Mip-NeRF360}
& \multicolumn{4}{c|}{Tanks\&Temples}
& \multicolumn{4}{c}{Deep Blending} \\
Method
& PSNR$\uparrow$ & SSIM$\uparrow$ & LPIPS$\downarrow$ & Num
& PSNR$\uparrow$ & SSIM$\uparrow$ & LPIPS$\downarrow$ & Num
& PSNR$\uparrow$ & SSIM$\uparrow$ & LPIPS$\downarrow$ & Num \\
\hline
Position-only~\cite{steep}
& $27.36$ & $0.798$ & $0.239$ & $2.0$
& $23.94$ & $0.847$ & $0.178$ & $1.1$
& $29.47$ & $0.896$ & $0.255$ & $1.1$ \\
Joint position-scale (Ours)
& $\mathbf{27.72}$ & $\mathbf{0.818}$ & $\mathbf{0.219}$ & $1.9$
& $\mathbf{24.20}$ & $\mathbf{0.853}$ & $\mathbf{0.174}$ & $1.0$
& $\mathbf{29.79}$ & $\mathbf{0.906}$ & $\mathbf{0.245}$ & $1.3$ \\
\hline
\end{tabular}%
}
\end{table}
% \noindent\textbf{Ablation study.}
% Tab.~\ref{tab:ablation} evaluates the contribution of each component on Mip-NeRF360~\cite{mipnerf360}.
% Starting from the SteepGS~\cite{steep}~(a), adding our joint position-scale splitting~(b) improves all metrics while slightly reducing the number of Gaussians. It show that incorporating scale into the loss-aware split direction yields more effective offspring placement.
% Introducing the cloning surrogate (c) further improves all metrics, and produces the most compact representation among the variants.
% This shows that surrogate-guided cloning avoids unnecessary duplication by allocating new primitives only when the duplicated contribution is predicted to reduce the loss.
% Finally, adding the opacity decay surrogate (d) gives the best overall reconstruction quality across all metrics.
% These results show that the three components play complementary roles: joint splitting improves local geometry refinement, cloning improves primitive allocation, and opacity decay stabilizes suppression of harmful contributions.

\noindent\textbf{Ablation study.}
Tab.~\ref{tab:ablation} evaluates the contribution of each component on Mip-NeRF360~\cite{mipnerf360}.
Starting from the SteepGS baseline~(a), adding joint position-scale splitting~(b), surrogate-guided cloning~(c), and surrogate-guided opacity decay~(d) progressively improves all reconstruction metrics.
In particular, the full model achieves the best PSNR, SSIM, and LPIPS among all variants.
These results show that all key components play complementary roles in improving loss-aware density control. 

To further isolate the contribution of joint position-scale splitting, we compare it with the position-only splitting of SteepGS~\cite{steep}, while keeping our surrogate-guided cloning, opacity decay, and all other training settings unchanged. As shown in Tab.~\ref{tab:split_comparison}, joint position-scale splitting consistently improves PSNR, SSIM, and LPIPS across all three benchmarks while using comparable numbers of Gaussians. These results demonstrate that jointly adapting offspring positions and scales improves reconstruction quality compared with position-only splitting.

\begin{figure}[t!]
\centering
\vspace{-2pt}
\begin{minipage}[t]{0.65\textwidth}
    \vspace{0pt}
    \centering
    \includegraphics[
        width=\linewidth,
        trim=0mm 6mm 6mm 0mm,
        clip
    ]{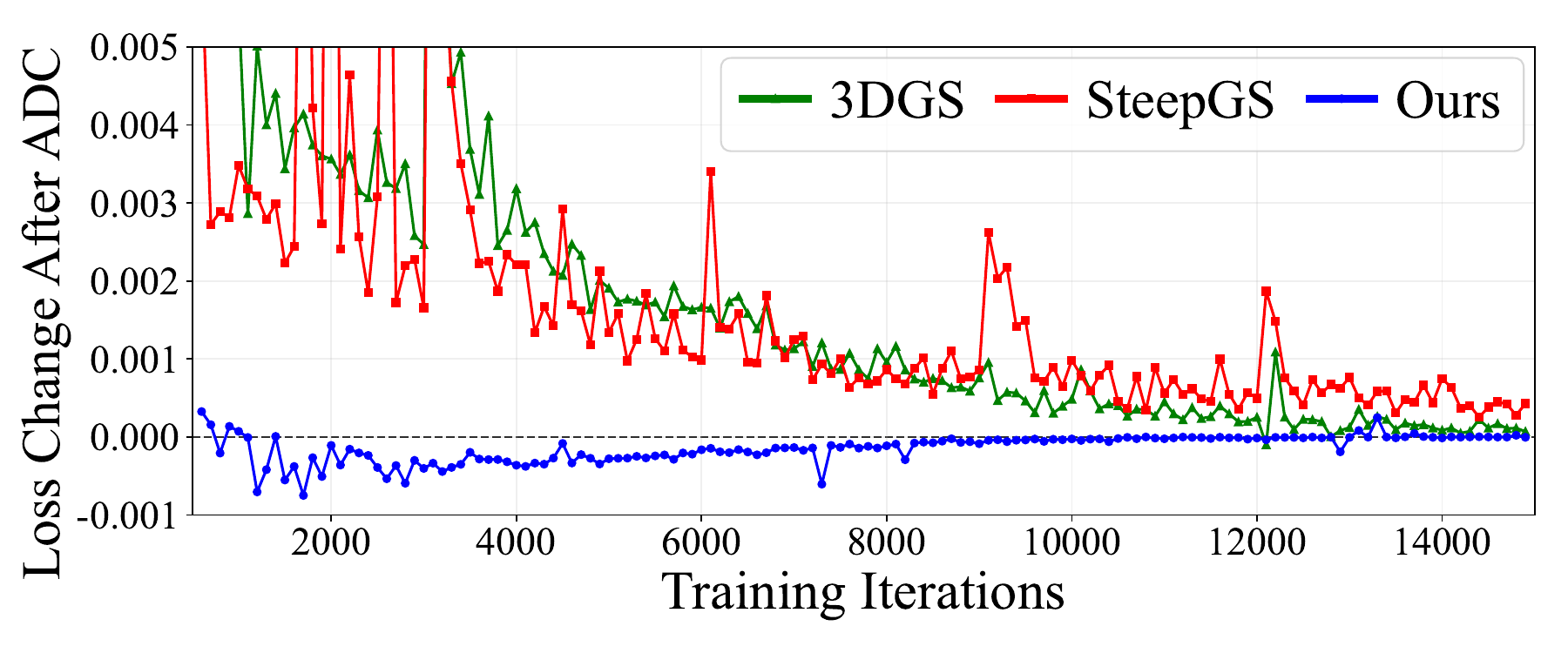}
    \vspace{-12pt}
    \caption{Loss change comparison after density-control.}
    \label{fig:lossdiff}
\end{minipage}
\hfill
\begin{minipage}[t]{0.34\textwidth}
    \vspace{2pt}
    \centering
    \includegraphics[
        width=\linewidth,
        trim=2.5mm 3.5mm 2.2mm 0mm,
        clip
    ]{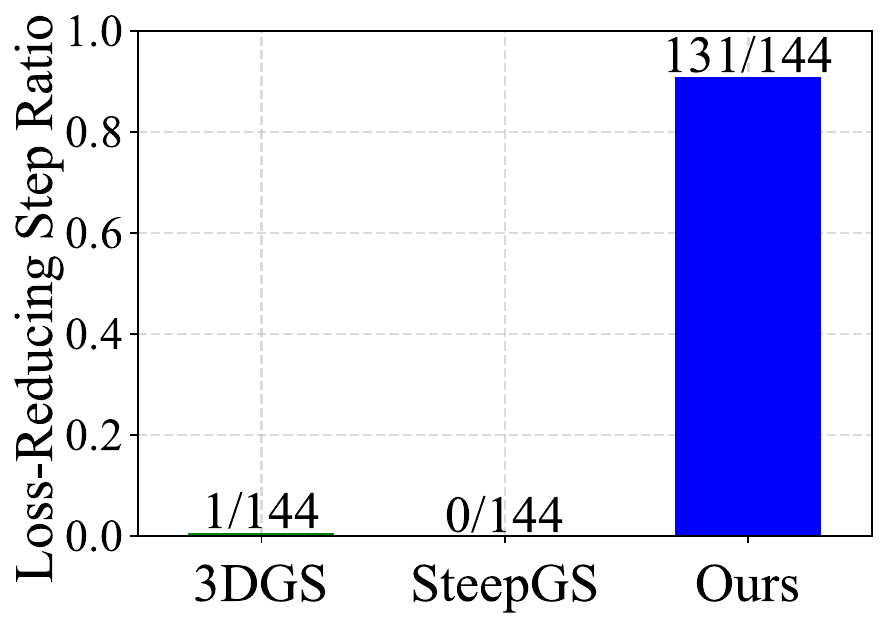}
    \vspace{-5pt}
    \captionof{figure}{Loss-reducing step ratio.}
    \label{fig:reduce_ratio}
\end{minipage}

\end{figure}

\noindent\textbf{Validation of the loss-reducing densification.}
To assess whether densification reduces the objective, we compute 
\(\Delta\mathcal{L}=\mathcal{L}_{\mathrm{after}}-\mathcal{L}_{\mathrm{before}}\) 
at each densification step, where negative values indicate loss reduction.
Each marker in Fig.~\ref{fig:lossdiff} denotes one densification step, and Fig.~\ref{fig:reduce_ratio} reports the fraction of steps with \(\Delta\mathcal{L}<0\).
We select a single viewpoint from the Counter scene of
Mip-NeRF360~\cite{mipnerf360} and keep it fixed across all
densification steps when measuring \(\Delta\mathcal{L}\).
As shown in Figs.~\ref{fig:lossdiff} and~\ref{fig:reduce_ratio}, our method yields negative loss changes for most densification steps, showing that the proposed operations update Gaussians toward loss-reducing directions.
In contrast, 3DGS does not consider the loss effect of densification, while SteepGS accounts for it only in splitting; thus, neither consistently reduces the loss across all density-control operations.
These results show that our loss-aware densification more effectively reduces the actual training loss than conventional ADC.

\begin{table}[t]
\caption{
Comparison of training cost and rendering speed (FPS) on a single
NVIDIA A6000 GPU.
}
\vspace{5pt}
\label{tab:computational_cost}
\centering
\small
\setlength{\tabcolsep}{4pt}
\renewcommand{\arraystretch}{1.0}
\begin{tabular}{@{}c|ccc@{}}
\hline
Method
& Training time (min)$\downarrow$
& Peak memory (GB)$\downarrow$
& FPS$\uparrow$ \\
\hline
3DGS~\cite{3dgs}
& $\mathbf{32.5}$
& $\mathbf{9.91}$
& $157$ \\
Ours
& $47.5$
& $11.32$
& $\mathbf{232}$ \\
\hline
\end{tabular}
\end{table}

\noindent\textbf{Computational cost.} Tab.~\ref{tab:computational_cost} compares our method with 3DGS~\cite{3dgs} in terms of training time, peak GPU memory usage during training, and rendering speed on a single NVIDIA A6000 GPU. Accumulating the joint position-scale splitting matrix and surrogate curvature statistics within the rasterizer increases training time and peak GPU memory usage. However, this overhead is limited to training; inference uses standard Gaussian rasterization without surrogate computation. Our more compact Gaussian representation enables faster rendering in this comparison, illustrating a trade-off between training cost and rendering efficiency.

\section{Conclusion}

% We presented a more principled adaptive density control framework for 3D Gaussian Splatting by reformulating split, clone, and opacity-reduction decisions through local surrogate-based criteria. Our method extends Hessian-guided splitting to the joint position-scale space, introduces a deterministic duplicate-cloning rule from a local quadratic surrogate, and replaces global opacity reset with surrogate-based soft opacity reduction, leading to better Gaussian allocation during training and improved rendering performance under comparable budgets.
We introduce a unified loss-aware density control framework for 3D Gaussian Splatting. Instead of relying on heuristic gradient, scale, and opacity reset rules, our method predicts the local loss effect of splitting, cloning, and opacity decay using efficient per-Gaussian surrogates. By combining joint position-scale splitting, surrogate-guided cloning, and opacity decay, the proposed method allocates primitives more effectively and suppresses harmful contributions more stably. Experiments across the benchmarks show that ours improves the quality-compactness trade-off with fewer Gaussians.

\section{Acknowledgements}

This work was supported in part by MSIT/IITP (No. RS-2022-II220680, RS-2020-II201821, RS-2019-II190421, RS-2024-00459618, RS-2024-00360227, RS-2026-25549037, RS-2025-25442569),  MSIT/NIPA (No. RS-2026-25621569), MSIT/NRF (No. RS-2024-00357729), and KNPA/KIPoT (No. RS-2025-25393280).

{
\small

\bibliographystyle{plainnat}
\bibliography{references}

@Article{3dgs,
      author       = {Kerbl, Bernhard and Kopanas, Georgios and Leimk{\"u}hler, Thomas and Drettakis, George},
      title        = {3D Gaussian Splatting for Real-Time Radiance Field Rendering},
      journal      = {ACM Transactions on Graphics},
      number       = {4},
      volume       = {42},
      month        = {July},
      year         = {2023},
}

@inproceedings{nvs,
  title={Novel view synthesis in tensor space},
  author={Avidan, Shai and Shashua, Amnon},
  booktitle={Proceedings of IEEE computer society conference on computer vision and pattern recognition},
  pages={1034--1040},
  year={1997},
  organization={IEEE}
}

@inproceedings{steep,
  title={Steepest descent density control for compact 3D Gaussian splatting},
  author={Wang, Peihao and Wang, Yuehao and Wang, Dilin and Mohan, Sreyas and Fan, Zhiwen and Wu, Lemeng and Cai, Ruisi and Yeh, Yu-Ying and Wang, Zhangyang and Liu, Qiang and others},
  booktitle={Proceedings of the Computer Vision and Pattern Recognition Conference},
  pages={26663--26672},
  year={2025}
}

@inproceedings{absgs,
  title={Absgs: Recovering fine details in 3d gaussian splatting},
  author={Ye, Zongxin and Li, Wenyu and Liu, Sidun and Qiao, Peng and Dou, Yong},
  booktitle={Proceedings of the 32nd ACM international conference on multimedia},
  pages={1053--1061},
  year={2024}
}

@article{mcmc,
  title={3d gaussian splatting as markov chain monte carlo},
  author={Kheradmand, Shakiba and Rebain, Daniel and Sharma, Gopal and Sun, Weiwei and Tseng, Yang-Che and Isack, Hossam and Kar, Abhishek and Tagliasacchi, Andrea and Yi, Kwang Moo},
  journal={Advances in Neural Information Processing Systems},
  volume={37},
  pages={80965--80986},
  year={2024}
}

@article{nerf,
  title={Nerf: Representing scenes as neural radiance fields for view synthesis},
  author={Mildenhall, Ben and Srinivasan, Pratul P and Tancik, Matthew and Barron, Jonathan T and Ramamoorthi, Ravi and Ng, Ren},
  journal={Communications of the ACM},
  volume={65},
  number={1},
  pages={99--106},
  year={2021},
  publisher={ACM New York, NY, USA}
}

@inproceedings{revising,
  title={Revising densification in gaussian splatting},
  author={Rota Bul{\`o}, Samuel and Porzi, Lorenzo and Kontschieder, Peter},
  booktitle={European Conference on Computer Vision},
  pages={347--362},
  year={2024},
  organization={Springer}
}

@inproceedings{GDAGS,
  title     = {Gradient-Direction-Aware Density Control for {3D} Gaussian Splatting},
  author    = {Zhou, Zheng and Xiong, Yu-Jie and Zhang, Jia-Chen and Xia, Chun-Ming and Qiu, Xihe and Zhan, Hongjian},
  booktitle = {International Conference on Learning Representations},
  year      = {2026}
}

@inproceedings{Lpips,
  title={The unreasonable effectiveness of deep features as a perceptual metric},
  author={Zhang, Richard and Isola, Phillip and Efros, Alexei A and Shechtman, Eli and Wang, Oliver},
  booktitle={Proceedings of the IEEE conference on computer vision and pattern recognition},
  pages={586--595},
  year={2018}
}

@inproceedings{mipnerf360,
  title={Mip-nerf 360: Unbounded anti-aliased neural radiance fields},
  author={Barron, Jonathan T and Mildenhall, Ben and Verbin, Dor and Srinivasan, Pratul P and Hedman, Peter},
  booktitle={Proceedings of the IEEE/CVF conference on computer vision and pattern recognition},
  pages={5470--5479},
  year={2022}
}

@article{Deepblending,
  title={Deep blending for free-viewpoint image-based rendering},
  author={Hedman, Peter and Philip, Julien and Price, True and Frahm, Jan-Michael and Drettakis, George and Brostow, Gabriel},
  journal={ACM Transactions on Graphics (ToG)},
  volume={37},
  number={6},
  pages={1--15},
  year={2018},
  publisher={ACM New York, NY, USA}
}

@article{tankandtemples,
  title={Tanks and temples: Benchmarking large-scale scene reconstruction},
  author={Knapitsch, Arno and Park, Jaesik and Zhou, Qian-Yi and Koltun, Vladlen},
  journal={ACM Transactions on Graphics (ToG)},
  volume={36},
  number={4},
  pages={1--13},
  year={2017},
  publisher={ACM New York, NY, USA}
}

@inproceedings{compact,
  title={Compact 3d gaussian representation for radiance field},
  author={Lee, Joo Chan and Rho, Daniel and Sun, Xiangyu and Ko, Jong Hwan and Park, Eunbyung},
  booktitle={Proceedings of the IEEE/CVF Conference on Computer Vision and Pattern Recognition},
  pages={21719--21728},
  year={2024}
}

@inproceedings{mh,
  title={Metropolis-Hastings Sampling for 3D Gaussian Reconstruction},
  author={Hyunjin Kim and Haebeom Jung and Jaesik Park},
  booktitle={The Thirty-ninth Annual Conference on Neural Information Processing Systems},
  year={2025}
}

@inproceedings{inverserendering,
  title={Gs-ir: 3d gaussian splatting for inverse rendering},
  author={Liang, Zhihao and Zhang, Qi and Feng, Ying and Shan, Ying and Jia, Kui},
  booktitle={Proceedings of the IEEE/CVF Conference on Computer Vision and Pattern Recognition},
  pages={21644--21653},
  year={2024}
}

@inproceedings{gaussiangrouping,
  title={Gaussian grouping: Segment and edit anything in 3d scenes},
  author={Ye, Mingqiao and Danelljan, Martin and Yu, Fisher and Ke, Lei},
  booktitle={European conference on computer vision},
  pages={162--179},
  year={2024},
  organization={Springer}
}

@inproceedings{diffusion_based_editing,
  title={Diffusion feature field for text-based 3d editing with gaussian splatting},
  author={Koh, Eunseo and Hyun, Sangeek and Lee, MinKyu and Chung, Jiwoo and Seo, Kangmin and Heo, Jae-Pil},
  booktitle={Advances in Neural Information Processing Systems},
  volume={38},
  pages={159--178},
  year={2025}
}

@inproceedings{resgs,
  title={Resgs: Residual densification of 3d gaussian for efficient detail recovery},
  author={Lyu, Yanzhe and Cheng, Kai and Kang, Xin and Chen, Xuejin},
  booktitle={Proceedings of the IEEE/CVF International Conference on Computer Vision},
  pages={28093--28102},
  year={2025}
}

@inproceedings{gsslam,
  title={Gaussian splatting slam},
  author={Matsuki, Hidenobu and Murai, Riku and Kelly, Paul HJ and Davison, Andrew J},
  booktitle={Proceedings of the IEEE/CVF conference on computer vision and pattern recognition},
  pages={18039--18048},
  year={2024}
}

@inproceedings{pixelgs,
  title={Pixel-gs: Density control with pixel-aware gradient for 3d gaussian splatting},
  author={Zhang, Zheng and Hu, Wenbo and Lao, Yixing and He, Tong and Zhao, Hengshuang},
  booktitle={European Conference on Computer Vision},
  pages={326--342},
  year={2024},
  organization={Springer}
}

@inproceedings{dc4gs,
    title       = {{DC4GS: Directional Consistency-Driven Adaptive Density Control for 3D Gaussian Splatting}},
    author      = {Jeong, Moonsoo and Kim, Dongbeen and Kim, Minseong and Lee, Sungkil},
    booktitle   = {Advances in Neural Information Processing Systems (NeurIPS)},
    year        = {2025}
}

@inproceedings{seamless,
  title={Towards realistic example-based modeling via 3d gaussian stitching},
  author={Gao, Xinyu and Yang, Ziyi and Gong, Bingchen and Han, Xiaoguang and Yang, Sipeng and Jin, Xiaogang},
  booktitle={Proceedings of the Computer Vision and Pattern Recognition Conference},
  pages={26597--26607},
  year={2025}
}

@inproceedings{zhang2024fregs,
  title={Fregs: 3d gaussian splatting with progressive frequency regularization},
  author={Zhang, Jiahui and Zhan, Fangneng and Xu, Muyu and Lu, Shijian and Xing, Eric},
  booktitle={Proceedings of the IEEE/CVF Conference on Computer Vision and Pattern Recognition},
  pages={21424--21433},
  year={2024}
}

@inproceedings{molchanov2016pruning,
  title     = {Pruning Convolutional Neural Networks for Resource Efficient Inference},
  author    = {Molchanov, Pavlo and Tyree, Stephen and Karras, Tero and Aila, Timo and Kautz, Jan},
  booktitle = {International Conference on Learning Representations},
  year      = {2017}
}

@article{lecun1989optimal,
  title={Optimal brain damage},
  author={LeCun, Yann and Denker, John and Solla, Sara},
  journal={Advances in neural information processing systems},
  volume={2},
  year={1989}
}

@article{wu2020firefly,
  title={Firefly neural architecture descent: a general approach for growing neural networks},
  author={Wu, Lemeng and Liu, Bo and Stone, Peter and Liu, Qiang},
  journal={Advances in neural information processing systems},
  volume={33},
  pages={22373--22383},
  year={2020}
}

@inproceedings{molchanov2019importance,
  title={Importance estimation for neural network pruning},
  author={Molchanov, Pavlo and Mallya, Arun and Tyree, Stephen and Frosio, Iuri and Kautz, Jan},
  booktitle={Proceedings of the IEEE/CVF conference on computer vision and pattern recognition},
  pages={11264--11272},
  year={2019}
}

@article{charbonnier1997deterministic,
  title={Deterministic edge-preserving regularization in computed imaging},
  author={Charbonnier, Pierre and Blanc-F{\'e}raud, Laure and Aubert, Gilles and Barlaud, Michel},
  journal={IEEE Transactions on image processing},
  volume={6},
  number={2},
  pages={298--311},
  year={1997},
  publisher={IEEE}
}

@inproceedings{barron2019general,
  title={A general and adaptive robust loss function},
  author={Barron, Jonathan T},
  booktitle={Proceedings of the IEEE/CVF conference on computer vision and pattern recognition},
  pages={4331--4339},
  year={2019}
}

@inproceedings{hollein20253dgs,
  title={3dgs-lm: Faster gaussian-splatting optimization with levenberg-marquardt},
  author={H{\"o}llein, Lukas and Bo{\v{z}}i{\v{c}}, Alja{\v{z}} and Zollh{\"o}fer, Michael and Nie{\ss}ner, Matthias},
  booktitle={Proceedings of the IEEE/CVF International Conference on Computer Vision},
  pages={26740--26750},
  year={2025}
}

@inproceedings{pehlivan2026matrix,
  title={Matrix-free Second-order Optimization of Gaussian Splats with Residual Sampling},
  author={Pehlivan, Hamza and Camiletto, Andrea Boscolo and Foo, Lin Geng and Habermann, Marc and Theobalt, Christian},
  booktitle={2026 International Conference on 3D Vision (3DV)},
  pages={18--28},
  year={2026},
  organization={IEEE}
}

@inproceedings{fang2024mini,
  title={Mini-splatting: Representing scenes with a constrained number of gaussians},
  author={Fang, Guangchi and Wang, Bing},
  booktitle={European conference on computer vision},
  pages={165--181},
  year={2024},
  organization={Springer}
}

@article{dongreframing,
  title={Reframing Gaussian splatting densification with complexity-density consistency of primitives},
  author={Dong, Zhemeng and Jiang, Junjun and Chen, Youyu and Zhang, Jiaxin and Jiang, Kui and Liu, Xianming},
  journal={Advances in Neural Information Processing Systems},
  volume={38},
  pages={127410--127433},
  year={2025}
}

@article{ssim,
  title={Image quality assessment: from error visibility to structural similarity},
  author={Wang, Zhou and Bovik, Alan C and Sheikh, Hamid R and Simoncelli, Eero P},
  journal={IEEE transactions on image processing},
  volume={13},
  number={4},
  pages={600--612},
  year={2004},
  publisher={IEEE}
}

@inproceedings{mipsplatting,
  title={Mip-splatting: Alias-free 3d gaussian splatting},
  author={Yu, Zehao and Chen, Anpei and Huang, Binbin and Sattler, Torsten and Geiger, Andreas},
  booktitle={2024 IEEE/CVF Conference on Computer Vision and Pattern Recognition (CVPR)},
  pages={19447--19456},
  year={2024},
  organization={IEEE}
}

@inproceedings{2dgs,
  title={2d gaussian splatting for geometrically accurate radiance fields},
  author={Huang, Binbin and Yu, Zehao and Chen, Anpei and Geiger, Andreas and Gao, Shenghua},
  booktitle={ACM SIGGRAPH 2024 conference papers},
  pages={1--11},
  year={2024}
}

@article{hierarchicalgs,
  title={A hierarchical 3d gaussian representation for real-time rendering of very large datasets},
  author={Kerbl, Bernhard and Meuleman, Andreas and Kopanas, Georgios and Wimmer, Michael and Lanvin, Alexandre and Drettakis, George},
  journal={ACM Transactions On Graphics (TOG)},
  volume={43},
  number={4},
  pages={1--15},
  year={2024},
  publisher={ACM, New York, NY, USA}
}

@inproceedings{ever,
  title={Ever: Exact volumetric ellipsoid rendering for real-time view synthesis},
  author={Mai, Alexander and Hedman, Peter and Kopanas, George and Verbin, Dor and Futschik, David and Xu, Qiangeng and Kuester, Falko and Barron, Jonathan T and Zhang, Yinda},
  booktitle={2025 IEEE/CVF International Conference on Computer Vision (ICCV)},
  pages={4930--4939},
  year={2025},
  organization={IEEE}
}

@inproceedings{betasplatting,
  title={Deformable beta splatting},
  author={Liu, Rong and Sun, Dylan and Chen, Meida and Wang, Yue and Feng, Andrew},
  booktitle={Proceedings of the special interest group on computer graphics and interactive techniques conference conference papers},
  pages={1--11},
  year={2025}
}

@inproceedings{multigs,
  title={Multi-scale 3d gaussian splatting for anti-aliased rendering},
  author={Yan, Zhiwen and Low, Weng Fei and Chen, Yu and Lee, Gim Hee},
  booktitle={2024 IEEE/CVF Conference on Computer Vision and Pattern Recognition (CVPR)},
  pages={20923--20931},
  year={2024},
  organization={IEEE}
}

@inproceedings{vol3d,
  title={Volumetrically consistent 3d gaussian rasterization},
  author={Talegaonkar, Chinmay and Belhe, Yash and Ramamoorthi, Ravi and Antipa, Nicholas},
  booktitle={2025 IEEE/CVF Conference on Computer Vision and Pattern Recognition (CVPR)},
  pages={10953--10963},
  year={2025},
  organization={IEEE}
}

@inproceedings{sadgs,
  title     = {Faster {3D} Gaussian Splatting Convergence via Structure-Aware Densification},
  author    = {Lyu, Linjie and Tewari, Ayush and Chen, Jianchun and Leimk{\"u}hler, Thomas and Theobalt, Christian},
  booktitle = {ACM SIGGRAPH 2026 Conference Papers},
  year      = {2026},
  doi       = {10.1145/3799902.3811212}
}
}

%%%%%%%%%%%%%%%%%%%%%%%%%%%%%%%%%%%%%%%%%%%%%%%%%%%%%%%%%%%%
\newpage
\FloatBarrier
\appendix

\section*{\centering Appendix}

\section{More implementation detail}

\subsection{Experiment Setting and Hyperparameters}
\label{app:hyperparameters}

We adopt the same training configuration as the original 3DGS, including learning rates, the optimization schedule, and the total number of training iterations. For density control, we follow the default 3DGS protocol: densification starts from iteration 500, is performed every 100 iterations, and is applied until iteration 15,000. The opacity threshold for pruning is also kept the same as in the original 3DGS.

All hyperparameters are fixed for every experiment across all scenes and benchmarks. For the reproduced baselines, we follow the official configurations provided by their released codebases. However, since some official implementations enable exposure compensation while others do not, we disable exposure compensation for all reproduced baselines and our method to ensure a fair comparison. For the 3DGS results in the main Tab.~\ref{tab:main_results}, we use the officially released pretrained models provided by 3DGS. 

For all experiments, we empirically set the hyperparameters used in our method to \(\tau_{\mathrm{pos}}=\tau_{\mathrm{scale}}=10^{-4}\),
which serve as thresholds for determining whether a Gaussian primitive is at a stationary point, and 
\(\tau_{\mathrm{split}}=-10^{-8}\), 
\(\tau_{\mathrm{clone}}=10^{-6}\), and 
\(\tau_{\mathrm{decay}}=10^{-4}\), 
which are the main hyperparameters that significantly affect loss-aware densification.  We also report a sensitivity analysis for the main hyperparameters in the densification process of our method, 
$\tau_{\mathrm{split}}$, $\tau_{\mathrm{clone}}$, and $\tau_{\mathrm{decay}}$, 
as shown in Tables~\ref{app:split_hyper}, \ref{app:clone_hyper}, and \ref{app:decay_hyper}.

\begin{table}[t]
\caption{Experiment on the splitting hyperparameter. We set $\tau_{\mathrm{split}}=-10^{-8}$ for default configuration.}
\centering
\scriptsize
\setlength{\tabcolsep}{3pt}
\renewcommand{\arraystretch}{1.5}
\resizebox{\columnwidth}{!}{%
\begin{tabular}{c|cccc|cccc|cccc}
\hline
Dataset
& \multicolumn{4}{c|}{Mip-NeRF360}
& \multicolumn{4}{c|}{Tanks\&Temples}
& \multicolumn{4}{c}{Deep Blending} \\
Method
& PSNR$\uparrow$ & SSIM$\uparrow$ & LPIPS$\downarrow$ & Num
& PSNR$\uparrow$ & SSIM$\uparrow$ & LPIPS$\downarrow$ & Num
& PSNR$\uparrow$ & SSIM$\uparrow$ & LPIPS$\downarrow$ & Num \\
\hline
Ours~($\tau_{\mathrm{split}}=-10^{-7}$)
& $27.68$ & $0.815$ & $0.225$ & $1.6$
& $24.11$ & $0.852$ & $0.176$ & $0.8$
& $29.61$ & $0.905$ & $0.245$ & $1.2$ \\

\underline{Ours~($\tau_{\mathrm{split}}=-10^{-8}$)}
& $27.72$ & $0.818$ & $0.219$ & $1.9$
& $24.20$ & $0.853$ & $0.174$ & $1.0$
& $29.79$ & $0.906$ & $0.245$ & $1.3$ \\

Ours~($\tau_{\mathrm{split}}=-10^{-9}$)
& $27.72$ & $0.818$ & $0.219$ & $2.0$
& $24.26$ & $0.853$ & $0.174$ & $1.1$
& $29.72$ & $0.906$ & $0.245$ & $1.3$ \\
\hline
\end{tabular}
}
\label{app:split_hyper}
\end{table}

\begin{table}[t]
\caption{Experiment on the clone hyperparameter. We set $\tau_{\mathrm{clone}}=10^{-6}$ for default configuration.}
\centering
\scriptsize
\setlength{\tabcolsep}{3pt}
\renewcommand{\arraystretch}{1.5}
\resizebox{\columnwidth}{!}{%
\begin{tabular}{c|cccc|cccc|cccc}
\hline
Dataset
& \multicolumn{4}{c|}{Mip-NeRF360}
& \multicolumn{4}{c|}{Tanks\&Temples}
& \multicolumn{4}{c}{Deep Blending} \\
Method
& PSNR$\uparrow$ & SSIM$\uparrow$ & LPIPS$\downarrow$ & Num
& PSNR$\uparrow$ & SSIM$\uparrow$ & LPIPS$\downarrow$ & Num
& PSNR$\uparrow$ & SSIM$\uparrow$ & LPIPS$\downarrow$ & Num \\
\hline
Ours~($\tau_{\mathrm{clone}}=10^{-5}$)
& $27.70$ & $0.818$ & $0.218$ & $1.9$
& $24.21$ & $0.854$ & $0.174$ & $1.0$
& $29.79$ & $0.905$ & $0.245$ & $1.3$ \\

\underline{Ours~($\tau_{\mathrm{clone}}=10^{-6}$})
& $27.72$ & $0.818$ & $0.219$ & $1.9$
& $24.20$ & $0.853$ & $0.174$ & $1.0$
& $29.79$ & $0.906$ & $0.245$ & $1.3$ \\

Ours~($\tau_{\mathrm{clone}}=10^{-7}$)
& $27.71$ & $0.818$ & $0.219$ & $1.9$
& $24.10$ & $0.853$ & $0.174$ & $1.0$
& $29.80$ & $0.906$ & $0.245$ & $1.4$ \\
\hline
\end{tabular}
}
\label{app:clone_hyper}
\end{table}

\begin{table}[t]
\caption{Experiment on the decay hyperparameter. We set $\tau_{\mathrm{decay}}=10^{-4}$ for default configuration.}
\centering
\scriptsize
\setlength{\tabcolsep}{3pt}
\renewcommand{\arraystretch}{1.5}
\resizebox{\columnwidth}{!}{%
\begin{tabular}{c|cccc|cccc|cccc}
\hline
Dataset
& \multicolumn{4}{c|}{Mip-NeRF360}
& \multicolumn{4}{c|}{Tanks\&Temples}
& \multicolumn{4}{c}{Deep Blending} \\
Method
& PSNR$\uparrow$ & SSIM$\uparrow$ & LPIPS$\downarrow$ & Num
& PSNR$\uparrow$ & SSIM$\uparrow$ & LPIPS$\downarrow$ & Num
& PSNR$\uparrow$ & SSIM$\uparrow$ & LPIPS$\downarrow$ & Num \\
\hline
Ours~($\tau_{\mathrm{decay}}=10^{-3}$)
& $27.71$ & $0.818$ & $0.219$ & $1.6$
& $24.00$ & $0.850$ & $0.184$ & $0.8$
& $29.51$ & $0.904$ & $0.250$ & $1.0$ \\

\underline{Ours~($\tau_{\mathrm{decay}}=10^{-4}$)}
& $27.72$ & $0.818$ & $0.219$ & $1.9$
& $24.20$ & $0.853$ & $0.174$ & $1.0$
& $29.79$ & $0.906$ & $0.245$ & $1.3$ \\

Ours~($\tau_{\mathrm{decay}}=10^{-5}$)
& $27.74$ & $0.818$ & $0.218$ & $2.0$
& $24.20$ & $0.854$ & $0.170$ & $1.2$
& $29.61$ & $0.904$ & $0.245$ & $1.5$ \\
\hline
\end{tabular}
}
\label{app:decay_hyper}
\end{table}
% \begin{table}[t]
% \caption{Experiment on the decay hyperparameter. We set $\tau_{\mathrm{dec}}=10^{-4}$ for all experiments.}
% \centering
% \scriptsize
% \setlength{\tabcolsep}{3pt}
% \renewcommand{\arraystretch}{1.1}
% \resizebox{\columnwidth}{!}{%
% \begin{tabular}{c|cccc|cccc|cccc}
% \hline
% Dataset
% & \multicolumn{4}{c|}{Mip-NeRF360}
% & \multicolumn{4}{c|}{Tanks\&Temples}
% & \multicolumn{4}{c}{Deep Blending} \\
% Method
% & PSNR$\uparrow$ & SSIM$\uparrow$ & LPIPS$\downarrow$ & Num
% & PSNR$\uparrow$ & SSIM$\uparrow$ & LPIPS$\downarrow$ & Num
% & PSNR$\uparrow$ & SSIM$\uparrow$ & LPIPS$\downarrow$ & Num \\
% \hline
% Ours
% & $27.72$ & $0.818$ & $0.219$ & $1.9$
% & $24.20$ & $0.853$ & $0.174$ & $1.0$
% & $29.79$ & $0.906$ & $0.245$ & $1.3$ \\

% Ours($\tau_{decay} = 10^{-3}$)
% & $27.71$ & $0.818$ & $0.219$ & $1.6$
% & $24.00$ & $0.850$ & $0.184$ & $0.8$
% & $29.51$ & $0.904$ & $0.250$ & $1.0$ \\

% Ours($\tau_{decay} = 10^{-5}$)
% & $27.74$ & $0.818$ & $0.218$ & $2.0$
% & $24.20$ & $0.854$ & $0.170$ & $1.2$
% & $29.61$ & $0.904$ & $0.245$ & $1.5$ \\
% \hline
% \end{tabular}
% }
% \label{app:decay_hyper}
% \end{table}

\subsection{Pseudo code}
Algorithm~\ref{alg:loss_aware_density_control} summarizes our loss-aware
density control. Instead of relying on the heuristic ADC rules, we make
per-Gaussian decisions using local loss surrogates for three operations: split,
clone, and opacity decay. During training, we accumulate the joint
position--scale splitting matrix together with the quadratic surrogate terms for
cloning and opacity decay. At each density-control interval, these
statistics are averaged and used to select split candidates, append beneficial
duplicate children, and suppress harmful opacity contributions. 

The three operations are applied sequentially, with splitting taking priority
over cloning and opacity decay. This ordering follows the strength of the underlying criteria. The split
decision is directly tied to the second-order splitting matrix and its minimum-eigenvalue direction, giving a principled descent direction for escaping stationary regions. In contrast, cloning and opacity decay are then applied to the remaining Gaussians, as
they correspond to small opacity-contribution perturbations estimated with proxy-curvature quadratic surrogates.  We therefore prioritize splitting and use the other two operations as residual actions for Gaussians that are not selected for
geometry-aware refinement. Since all statistics and decisions are defined independently for each Gaussian, the procedure is naturally parallelizable. As in SteepGS~\cite{steep}, the statistics used by our method are computed within the rasterizer.

\begin{algorithm}[t]
\caption{Loss-Aware Density Control}
\label{alg:loss_aware_density_control}
\begin{algorithmic}[1]
\Input Initial Gaussians $\theta=\{\theta^{(i)}\}_{i=1}^{n}$; loss function $\mathcal L$; total iterations $T$; ADC interval $T_{\mathrm{dc}}$.
\For{each training step $t=1,\cdots,T$}
    \State Sample a training view and update $\theta$ by standard gradient descent.
    \State Accumulate per-Gaussian loss-aware statistics for splitting, cloning, and opacity decay.

    \If{$t \bmod T_{\mathrm{dc}} = 0$}
        \Statex \textbf{Split.}
        \State Compute the normalized joint splitting matrix $\tilde{\mathcal S}_{\phi}^{(i)}$.
        \State Compute its minimum eigenpair for each Gaussian.
        \State Select split candidates $\mathcal I_{\mathrm{split}}$ using Eq.~\eqref{eq:split_candidate}.
        \State Split each $i\in\mathcal I_{\mathrm{split}}$ along the joint position--scale direction using Eq.~\eqref{eq:split_update_joint}.

        \Statex \textbf{Clone.}
        \State Compute the duplicate strength $\eta_{\mathrm{dup}}^{(i)}=-g_{\mathrm{dup}}^{(i)}/h_{\mathrm{dup}}^{(i)}$.
        \State Select clone candidates $\mathcal I_{\mathrm{clone}}$ using Eq.~\eqref{eq:clone_set_update}.
        \State Append duplicate children $\mathcal C(\theta;\eta_{\mathrm{dup}}^{(i)})$ for all $i\in\mathcal I_{\mathrm{clone}}$.

        \Statex \textbf{Opacity decay.}
        \State Compute the opacity decay strength $\eta_{\mathrm{dec}}^{(i)}=g_{\mathrm{dec}}^{(i)}/h_{\mathrm{dec}}^{(i)}$.
        \State Select opacity decay candidates $\mathcal I_{\mathrm{decay}}$ using Eq.~\eqref{eq:soft_prune_set_update}.
        \State Apply $\mathcal P(\theta;\eta_{\mathrm{dec}}^{(i)})$ for all $i\in\mathcal I_{\mathrm{decay}}$.

        \State Remove low-opacity Gaussians and reset all density-control accumulators.
    \EndIf
\EndFor
\State \Return Optimized Gaussians $\theta$.
\end{algorithmic}
\end{algorithm}

\noindent\textbf{Operation priority.}
To examine the effect of the priority order, we evaluate all six
permutations of splitting (S), cloning (C), and opacity decay (D).
We use Bicycle from Mip-NeRF360, Truck from Tanks\&Temples, and
Dr.~Johnson from Deep Blending, changing only the priority order
while keeping all other settings unchanged.
Tab.~\ref{tab:operation_priority} reports the average results over
these three scenes.
The adopted Split$\rightarrow$Clone$\rightarrow$Decay order achieves the highest
average PSNR and tied-best SSIM and LPIPS, providing empirical
support for this ordering.

\noindent\textbf{Adaptive arbitration.}
We also evaluate an adaptive loss-scale arbitration strategy that
selects, for each Gaussian, the applicable operation with the largest
predicted loss reduction.
Under the same evaluation setting, our fixed-priority strategy
achieves higher PSNR and SSIM and lower LPIPS than this variant
(Tab.~\ref{tab:adaptive_arbitration}).
A possible explanation is the different parameterizations and
approximation scales of the operation-specific surrogates.
Splitting is evaluated in the normalized joint position-scale space,
whereas cloning and opacity decay use one-dimensional Taylor
surrogates over opacity contributions.
Thus, although each surrogate estimates the local loss effect of its
corresponding operation, their raw values are not necessarily
calibrated for direct cross-operation comparison.
Calibrating these estimates for adaptive operation selection remains
a promising direction for future work.

\begin{table}[t]
\centering
\caption{
Effect of operation priority, averaged over Bicycle, Truck, and
Dr.~Johnson.
S, C, and D denote splitting, cloning, and opacity decay, respectively.}
\label{tab:operation_priority}
\small
\setlength{\tabcolsep}{6pt}
\renewcommand{\arraystretch}{1.1}
\begin{tabular}{c|cccc}
\toprule
Priority order
& PSNR$\uparrow$
& SSIM$\uparrow$
& LPIPS$\downarrow$
& Num \\
\midrule
C$\rightarrow$D$\rightarrow$S
& $26.92$ & $0.854$ & $0.201$ & $1.9$ \\
C$\rightarrow$S$\rightarrow$D
& $26.84$ & $0.854$ & $0.201$ & $1.9$ \\
D$\rightarrow$C$\rightarrow$S
& $26.69$ & $0.853$ & $0.204$ & $1.9$ \\
D$\rightarrow$S$\rightarrow$C
& $26.86$ & $0.854$ & $0.200$ & $2.1$ \\
S$\rightarrow$D$\rightarrow$C
& $26.76$ & $0.853$ & $0.201$ & $2.1$ \\
\midrule
S$\rightarrow$C$\rightarrow$D (Ours)
& $\mathbf{26.93}$
& $\mathbf{0.854}$
& $\mathbf{0.200}$
& $2.0$ \\
\bottomrule
\end{tabular}
\end{table}

\begin{table}[t]
\centering
\caption{
Comparison of adaptive arbitration and fixed priority on the same
three scenes as in Tab.~\ref{tab:operation_priority}.
}
\label{tab:adaptive_arbitration}
\small
\setlength{\tabcolsep}{5pt}
\renewcommand{\arraystretch}{1.1}
\begin{tabular}{c|cccc}
\toprule
Arbitration strategy
& PSNR$\uparrow$
& SSIM$\uparrow$
& LPIPS$\downarrow$
& Num \\
\midrule
Adaptive loss-scale arbitration
& $26.87$ & $0.853$ & $0.203$ & $2.0$ \\
Fixed S$\rightarrow$C$\rightarrow$D (Ours)
& $\mathbf{26.93}$
& $\mathbf{0.854}$
& $\mathbf{0.200}$
& $2.0$ \\
\bottomrule
\end{tabular}
\end{table}

%%%%%%%%%%%%%%%%%%%%%%%%%%%%%

\subsection{Splitting Matrix of SteepGS}
\label{app:steepSplitting}

We review the splitting matrix of SteepGS~\cite{steep} using the
notation introduced in Sec.~\ref{sec:method_split}. Let
\(\Phi=\{\phi^{(i)}\}_{i=1}^{n}\) denote the collection of local split variables.
In the original SteepGS implementation, this variable is chosen as the Gaussian
center, i.e., \(\phi^{(i)}=p^{(i)}\). More generally, the same derivation can be
applied to an extended local variable, such as the joint position-scale variable
used in Sec.~\ref{sec:method_split}, as long as the projected Gaussian
contribution is twice differentiable with respect to that variable.

For a camera--pixel pair \((\varphi,x)\), we view the per-pixel objective as a
function of the projected Gaussian contributions:
\begin{equation}
\mathcal L(\Phi)
=
\mathbb{E}_{(\varphi,x)}
\left[
\mathcal L_{\varphi,x}
\bigl(
\alpha^{(1)}(x;\phi^{(1)}),\ldots,
\alpha^{(n)}(x;\phi^{(n)})
\bigr)
\right].
\label{eq:app_loss_alpha_functional}
\end{equation}
Here, \(\alpha^{(i)}(x;\phi^{(i)})\) denotes the image-space contribution of
Gaussian \(i\), corresponding to \(\sigma_\varphi(x;\theta^{(i)})\) in
SteepGS~\cite{steep}. The alpha-compositing operation, color attributes, and
the comparison with the ground-truth pixel are absorbed into
\(\mathcal L_{\varphi,x}\). This abstraction allows us to analyze the local
effect of splitting through the scalar contribution of each Gaussian. For the
Taylor analysis below, we assume that the resulting objective is differentiable
in the chosen local split variables.

We consider splitting Gaussian \(i\) into \(m_i\) offspring. The local split
variable of the \(j\)-th offspring is denoted by \(\phi_j^{(i)}\), and its
opacity weight is denoted by \(w_j^{(i)}\ge 0\). To preserve the local opacity
mass, the weights satisfy
\begin{equation}
\sum_{j=1}^{m_i} w_j^{(i)} = 1.
\label{eq:app_weight_constraint}
\end{equation}
Under this constraint, the parent contribution is locally replaced by a
weighted sum of offspring contributions:
\begin{equation}
\alpha^{(i)}(x;\phi^{(i)})
\quad\longrightarrow\quad
\sum_{j=1}^{m_i}
w_j^{(i)}
\alpha^{(i)}(x;\phi_j^{(i)}).
\label{eq:app_split_replacement}
\end{equation}
Therefore, the weight \(w_j^{(i)}\) is absorbed into the opacity of the
\(j\)-th offspring after splitting. Following SteepGS, we decompose each
offspring displacement into a shared mean shift and an individual zero-mean
splitting offset:
\begin{equation}
\phi_j^{(i)}
=
\phi^{(i)}
+
\varepsilon
\bigl(
\mu^{(i)}+\delta_j^{(i)}
\bigr),
\qquad
\sum_{j=1}^{m_i}
w_j^{(i)}
\delta_j^{(i)}
=
0.
\label{eq:app_split_decomposition}
\end{equation}
The scalar \(\varepsilon>0\) controls the perturbation size. The vector
\(\mu^{(i)}\) is the weighted mean displacement of the offspring group, while
\(\delta_j^{(i)}\) describes the relative displacement of each offspring around
this mean.

We first consider the effect of splitting a single Gaussian \(i\), while all
other Gaussians are fixed. Define the first-order term
\begin{equation}
g_{\phi}^{(i)}
=
\mathbb{E}_{(\varphi,x)}
\left[
\frac{\partial \mathcal L_{\varphi,x}}
{\partial \alpha^{(i)}(x)}
\nabla_{\phi^{(i)}}\alpha^{(i)}(x;\phi^{(i)})
\right].
\label{eq:app_first_order_alpha}
\end{equation}
By the chain rule, the Hessian of the loss with respect to \(\phi^{(i)}\) can
be decomposed as
\begin{equation}
\nabla_{\phi^{(i)}}^2 \mathcal L(\Phi)
=
\mathcal T_{\phi}^{(i)}(\theta)
+
\mathcal S_{\phi}^{(i)}(\theta),
\label{eq:app_hessian_decomp}
\end{equation}
where
\begin{equation}
\mathcal T_{\phi}^{(i)}(\theta)
=
\mathbb{E}_{(\varphi,x)}
\left[
\frac{\partial^2 \mathcal L_{\varphi,x}}
{\partial \alpha^{(i)}(x)^2}
\nabla_{\phi^{(i)}}\alpha^{(i)}(x;\phi^{(i)})
\nabla_{\phi^{(i)}}\alpha^{(i)}(x;\phi^{(i)})^{\top}
\right],
\label{eq:app_T_matrix}
\end{equation}
and
\begin{equation}
\mathcal S_{\phi}^{(i)}(\theta)
=
\mathbb{E}_{(\varphi,x)}
\left[
\frac{\partial \mathcal L_{\varphi,x}}
{\partial \alpha^{(i)}(x)}
\nabla_{\phi^{(i)}}^2\alpha^{(i)}(x;\phi^{(i)})
\right].
\label{eq:app_steepgs_split_matrix}
\end{equation}
The matrix \(\mathcal T_{\phi}^{(i)}\) comes from the curvature of the loss with
respect to the Gaussian contribution, whereas \(\mathcal S_{\phi}^{(i)}\) comes
from the curvature of the Gaussian contribution itself. Although both terms
appear in the Hessian of the ordinary parameter update, they play different
roles under splitting.

Using a Taylor expansion around \(\varepsilon=0\), the loss after splitting
Gaussian \(i\) becomes
\begin{align}
\mathcal L_i(\varepsilon)
=
\mathcal L(\Phi)
&+
\varepsilon
g_{\phi}^{(i)\top}
\sum_{j=1}^{m_i}
w_j^{(i)}
\bigl(
\mu^{(i)}+\delta_j^{(i)}
\bigr)
\nonumber\\
&+
\frac{\varepsilon^2}{2}
\left(
\sum_{j=1}^{m_i}
w_j^{(i)}
\bigl(
\mu^{(i)}+\delta_j^{(i)}
\bigr)
\right)^{\top}
\mathcal T_{\phi}^{(i)}
\left(
\sum_{j=1}^{m_i}
w_j^{(i)}
\bigl(
\mu^{(i)}+\delta_j^{(i)}
\bigr)
\right)
\nonumber\\
&+
\frac{\varepsilon^2}{2}
\sum_{j=1}^{m_i}
w_j^{(i)}
\bigl(
\mu^{(i)}+\delta_j^{(i)}
\bigr)^{\top}
\mathcal S_{\phi}^{(i)}
\bigl(
\mu^{(i)}+\delta_j^{(i)}
\bigr)
+
O(\varepsilon^3).
\label{eq:app_single_split_taylor}
\end{align}
By the zero-mean constraint
\(\sum_j w_j^{(i)}\delta_j^{(i)}=0\), the first-order term reduces to
\(g_{\phi}^{(i)\top}\mu^{(i)}\), and the
\(\mathcal T_{\phi}^{(i)}\) term depends only on the mean shift
\(\mu^{(i)}\). In contrast, the \(\mathcal S_{\phi}^{(i)}\) term separates as
\begin{equation}
\sum_{j=1}^{m_i}
w_j^{(i)}
\bigl(
\mu^{(i)}+\delta_j^{(i)}
\bigr)^{\top}
\mathcal S_{\phi}^{(i)}
\bigl(
\mu^{(i)}+\delta_j^{(i)}
\bigr)
=
\mu^{(i)\top}\mathcal S_{\phi}^{(i)}\mu^{(i)}
+
\sum_{j=1}^{m_i}
w_j^{(i)}
\delta_j^{(i)\top}
\mathcal S_{\phi}^{(i)}
\delta_j^{(i)}.
\label{eq:app_S_decomposition}
\end{equation}
Therefore,
\begin{equation}
\mathcal L_i(\varepsilon)
=
\mathcal L(\Phi)
+
\varepsilon g_{\phi}^{(i)\top}\mu^{(i)}
+
\frac{\varepsilon^2}{2}
\mu^{(i)\top}
\nabla_{\phi^{(i)}}^2 \mathcal L(\Phi)
\mu^{(i)}
+
\frac{\varepsilon^2}{2}
\sum_{j=1}^{m_i}
w_j^{(i)}
\delta_j^{(i)\top}
\mathcal S_{\phi}^{(i)}(\theta)
\delta_j^{(i)}
+
O(\varepsilon^3).
\label{eq:app_single_split_result}
\end{equation}

The first three terms describe the effect of moving the weighted average of the
offspring group. This behavior is equivalent to moving the original Gaussian
and can be handled by ordinary gradient-based optimization. The last term
depends only on the zero-mean offsets of the offspring and therefore captures
the intrinsic effect of splitting. This motivates the splitting characteristic
function
\begin{equation}
\Delta_{\mathrm{split}}^{(i)}
\bigl(
\{\delta_j^{(i)}\}_{j=1}^{m_i},
w^{(i)}
\bigr)
=
\frac{1}{2}
\sum_{j=1}^{m_i}
w_j^{(i)}
\delta_j^{(i)\top}
\mathcal S_{\phi}^{(i)}(\theta)
\delta_j^{(i)}.
\label{eq:app_split_characteristic}
\end{equation}
For this reason, \(\mathcal S_{\phi}^{(i)}(\theta)\) is called the splitting
matrix.

Applying the same expansion to all Gaussians and concatenating the mean shifts
as
\[
\mu =
\begin{bmatrix}
\mu^{(1)\top} & \cdots & \mu^{(n)\top}
\end{bmatrix}^{\top},
\]
we obtain the SteepGS loss decomposition:
\begin{equation}
\mathcal L(\Phi_{\mathrm{child}},w)
=
\mathcal L(\Phi)
+
\varepsilon
\nabla_{\Phi}\mathcal L(\Phi)^{\top}\mu
+
\frac{\varepsilon^2}{2}
\mu^{\top}
\nabla_{\Phi}^{2}\mathcal L(\Phi)
\mu
+
\frac{\varepsilon^2}{2}
\sum_{i=1}^{n}
\sum_{j=1}^{m_i}
w_j^{(i)}
\delta_j^{(i)\top}
\mathcal S_{\phi}^{(i)}(\theta)
\delta_j^{(i)}
+
O(\varepsilon^3).
\label{eq:app_steepgs_theorem1}
\end{equation}

This decomposition shows that splitting can locally reduce the objective only
when the splitting matrix has a negative-curvature direction. In particular,
if
\begin{equation}
\lambda_{\mathrm{min}}^{(i)}
=
\lambda_{\min}
\bigl(
\mathcal S_{\phi}^{(i)}(\theta)
\bigr)
< 0,
\end{equation}
then choosing the offspring offsets along the minimum-eigenvalue eigenvector
yields a negative value of \(\Delta_{\mathrm{split}}^{(i)}\). Under the
constraint \(\|\delta_j^{(i)}\|_2\le 1\), the SteepGS solution is
\begin{equation}
m_i = 2,
\qquad
w_1^{(i)}=w_2^{(i)}=\frac{1}{2},
\qquad
\delta_1^{(i)\star}=v_{\mathrm{min}}^{(i)},
\qquad
\delta_2^{(i)\star}=-v_{\mathrm{min}}^{(i)},
\label{eq:app_steepgs_solution}
\end{equation}
where \(v_{\mathrm{min}}^{(i)}\) is the eigenvector associated with
\(\lambda_{\mathrm{min}}^{(i)}\). Thus, SteepGS splits a Gaussian only when its
splitting matrix is not positive semidefinite, places the two offspring in
opposite directions along the steepest negative-curvature direction, and
assigns half of the parent opacity to each offspring.

%%%%%%%%%%%%%%%%%%%%%%%%%%%%%
\subsection{Normalized Joint Position-Scale Splitting Matrix}
\label{app:joint_pos_scale_matrix}

\paragraph{Joint position--scale splitting matrix.}
Following the second-order splitting view~\cite{steep}, the local effect of
splitting a Gaussian is characterized by the loss-weighted curvature of its
projected contribution. We extend the local split variable from the position
alone to the joint position--log-scale parameter
\begin{equation}
\phi^{(i)}
=
\begin{bmatrix}
p^{(i)}\\
\tilde{s}^{(i)}
\end{bmatrix}
\in\mathbb{R}^{6},
\qquad
\tilde{s}^{(i)}=\log s^{(i)} ,
\label{eq:app_joint_phi}
\end{equation}
where \(p^{(i)}\in\mathbb{R}^{3}\) is the Gaussian center and
\(\tilde{s}^{(i)}\in\mathbb{R}^{3}\) is the log-scale parameter.

For a camera--pixel pair \((\varphi,x)\), we introduce the screen-space variable
\begin{equation}
z^{(i)}(\phi^{(i)};\varphi)
=
\begin{bmatrix}
u^{(i)}(\phi^{(i)};\varphi)\\
q^{(i)}(\phi^{(i)};\varphi)
\end{bmatrix}
=
\begin{bmatrix}
u_x^{(i)},u_y^{(i)},q_{xx}^{(i)},q_{xy}^{(i)},q_{yy}^{(i)}
\end{bmatrix}^{\top}
\in\mathbb{R}^{5}.
\label{eq:app_screen_z}
\end{equation}
Here, \(u^{(i)}=\varphi(p^{(i)})\in\mathbb{R}^{2}\) is the projected mean, and
\(q^{(i)}=(q_{xx}^{(i)},q_{xy}^{(i)},q_{yy}^{(i)})\in\mathbb{R}^{3}\)
parameterizes the conic matrix of the inverse projected covariance,
\begin{equation}
Q(q^{(i)})
=
\begin{bmatrix}
q_{xx}^{(i)} & q_{xy}^{(i)}\\
q_{xy}^{(i)} & q_{yy}^{(i)}
\end{bmatrix}.
\label{eq:app_conic_Q}
\end{equation}
Let \(d^{(i)}=u^{(i)}-x\). The projected Gaussian contribution is written as
\begin{equation}
\alpha^{(i)}(x;\phi^{(i)})
=
o^{(i)}\,G^{(i)}(z^{(i)}),
\qquad
G^{(i)}(z^{(i)})
=
\exp\!\left(
-\frac{1}{2}
d^{(i)\top}
Q(q^{(i)})
d^{(i)}
\right),
\label{eq:app_projected_alpha}
\end{equation}
where \(o^{(i)}\) is the opacity. Since \(o^{(i)}\) is fixed when computing the
splitting direction, it is absorbed into the derivatives of
\(\alpha^{(i)}\).

The screen-space variable \(z^{(i)}\) is useful because rasterization naturally
provides quantities associated with the projected mean \(u^{(i)}\) and conic
\(q^{(i)}\). For each sampled camera--pixel pair, define
\begin{equation}
g_z^{(i)}(\varphi,x)
=
\frac{\partial \mathcal L_{\varphi,x}}
{\partial \alpha^{(i)}(x)}
\nabla_{z^{(i)}}\alpha^{(i)}(x;\phi^{(i)}),
\qquad
\mathcal H_z^{(i)}(\varphi,x)
=
\frac{\partial \mathcal L_{\varphi,x}}
{\partial \alpha^{(i)}(x)}
\nabla_{z^{(i)}}^2\alpha^{(i)}(x;\phi^{(i)}).
\label{eq:app_per_sample_screen_grad_hess}
\end{equation}
These are the loss-weighted gradient and Hessian of the projected Gaussian
contribution with respect to the intermediate screen-space variable
\(z^{(i)}\).

The Jacobian of \(z^{(i)}\) with respect to the joint variable
\(\phi^{(i)}\) is
\begin{equation}
\begin{gathered}
J_{up}^{(i)}
=
\frac{\partial u^{(i)}}{\partial p^{(i)}},
\qquad
J_{qp}^{(i)}
=
\frac{\partial q^{(i)}}{\partial p^{(i)}},
\qquad
J_{q\tilde{s}}^{(i)}
=
\frac{\partial q^{(i)}}{\partial \tilde{s}^{(i)}},
\\[3pt]
\mathcal J_z^{(i)}(\varphi)
=
\frac{\partial z^{(i)}}{\partial \phi^{(i)}}
=
\begin{bmatrix}
J_{up}^{(i)} & 0\\
J_{qp}^{(i)} & J_{q\tilde{s}}^{(i)}
\end{bmatrix}.
\end{gathered}
\label{eq:app_Jz}
\end{equation}
The zero block appears because the projected mean \(u^{(i)}\) depends on the
Gaussian position but not directly on its scale. Then, applying the second-order chain rule to
\(\alpha^{(i)}(x;\phi^{(i)})=\alpha^{(i)}(z^{(i)}(\phi^{(i)};\varphi))\) gives
\begin{equation}
\nabla_{\phi^{(i)}}^{2}\alpha^{(i)}
=
\mathcal J_z^{(i)\top}
\nabla_{z^{(i)}}^{2}\alpha^{(i)}
\mathcal J_z^{(i)}
+
\sum_{m=1}^{5}
\frac{\partial \alpha^{(i)}}{\partial z_m^{(i)}}
\nabla_{\phi^{(i)}}^{2}z_m^{(i)} .
\label{eq:app_second_order_chain}
\end{equation}
Multiplying by
\(\partial\mathcal L_{\varphi,x}/\partial\alpha^{(i)}(x)\) and taking the
expectation over sampled camera--pixel pairs yields the joint splitting matrix
\begin{equation}
\mathcal S_{\phi}^{(i)}(\theta)
=
\mathbb{E}_{(\varphi,x)}
\left[
\mathcal J_z^{(i)}(\varphi)^{\top}
\mathcal H_z^{(i)}(\varphi,x)
\mathcal J_z^{(i)}(\varphi)
+
\sum_{m=1}^{5}
g_{z_m}^{(i)}(\varphi,x)
\nabla_{\phi^{(i)}}^{2}
z_m^{(i)}(\phi^{(i)};\varphi)
\right].
\label{eq:app_joint_splitting_exact}
\end{equation}
This expression is the direct joint-variable counterpart of the SteepGS
splitting matrix. The first term pulls back the screen-space loss-weighted
Hessian to the joint position--scale space, while the second term accounts for
the nonlinear dependence of the screen-space variables \(z^{(i)}\) on
\(\phi^{(i)}\).

In block form,
\begin{equation}
\mathcal S_{\phi}^{(i)}
=
\begin{bmatrix}
\mathcal S_{pp}^{(i)} & \mathcal S_{p\tilde{s}}^{(i)}\\
\mathcal S_{\tilde{s}p}^{(i)} & \mathcal S_{\tilde{s}\tilde{s}}^{(i)}
\end{bmatrix}.
\label{eq:app_joint_block}
\end{equation}
For a single camera--pixel pair, the corresponding position--position block is
\begin{align}
\mathcal S_{pp}^{(i)}
=&
J_{up}^{(i)\top}
\mathcal H_{uu}^{(i)}
J_{up}^{(i)}
+
J_{up}^{(i)\top}
\mathcal H_{uq}^{(i)}
J_{qp}^{(i)}
+
J_{qp}^{(i)\top}
\mathcal H_{qu}^{(i)}
J_{up}^{(i)}
+
J_{qp}^{(i)\top}
\mathcal H_{qq}^{(i)}
J_{qp}^{(i)}
\nonumber\\
&+
\sum_{k=1}^{2}
g_{u_k}^{(i)}
\nabla_{p^{(i)}}^2 u_k^{(i)}
+
\sum_{r=1}^{3}
g_{q_r}^{(i)}
\nabla_{p^{(i)}}^2 q_r^{(i)} .
\label{eq:app_Spp}
\end{align}
The final \(\mathcal S_{pp}^{(i)}\) is obtained by averaging this quantity over
sampled camera--pixel pairs. The two summations are the second-order chain-rule
corrections: the first accounts for the curvature of the perspective
projection \(u^{(i)}=\varphi(p^{(i)})\), and the second accounts for the
curvature of the conic parameters with respect to the 3D position.

Similarly, the position--scale block for a single sample is
\begin{align}
\mathcal S_{p\tilde{s}}^{(i)}
=&
J_{up}^{(i)\top}
\mathcal H_{uq}^{(i)}
J_{q\tilde{s}}^{(i)}
+
J_{qp}^{(i)\top}
\mathcal H_{qq}^{(i)}
J_{q\tilde{s}}^{(i)}
+
\sum_{r=1}^{3}
g_{q_r}^{(i)}
\nabla_{p^{(i)}\tilde{s}^{(i)}}^2 q_r^{(i)} .
\label{eq:app_Sps_exact}
\end{align}
In our implementation, we use the more stable first-order pullback form and
omit the mixed second-order conic curvature term:
\begin{equation}
\mathcal S_{p\tilde{s}}^{(i)}
\approx
J_{up}^{(i)\top}
\mathcal H_{uq}^{(i)}
J_{q\tilde{s}}^{(i)}
+
J_{qp}^{(i)\top}
\mathcal H_{qq}^{(i)}
J_{q\tilde{s}}^{(i)},
\qquad
\mathcal S_{\tilde{s}p}^{(i)}
=
\mathcal S_{p\tilde{s}}^{(i)\top}.
\label{eq:app_Sps_impl}
\end{equation}
The scale--scale block is
\begin{equation}
\mathcal S_{\tilde{s}\tilde{s}}^{(i)}
=
J_{q\tilde{s}}^{(i)\top}
\mathcal H_{qq}^{(i)}
J_{q\tilde{s}}^{(i)}
+
\sum_{r=1}^{3}
g_{q_r}^{(i)}
\nabla_{\tilde{s}^{(i)}}^2 q_r^{(i)} .
\label{eq:app_Sss}
\end{equation}

The off-diagonal block \(\mathcal S_{p\tilde{s}}^{(i)}\) distinguishes the
joint formulation from a position-only splitting matrix. The term
\(J_{up}^{(i)\top}\mathcal H_{uq}^{(i)}J_{q\tilde{s}}^{(i)}\) captures the
interaction between the projected mean change induced by position and the
conic change induced by scale. The term
\(J_{qp}^{(i)\top}\mathcal H_{qq}^{(i)}J_{q\tilde{s}}^{(i)}\) captures the
interaction between position-induced and scale-induced conic changes.
Therefore, the minimum-eigenvalue direction of
\(\mathcal S_{\phi}^{(i)}\) estimates a splitting direction that jointly
accounts for the Gaussian center and its anisotropic spatial extent.

\paragraph{Normalized joint coordinate.}
The matrix \(\mathcal S_{\phi}^{(i)}\) is expressed in the raw
position--log-scale parameter space. However, the two blocks of
\(\phi^{(i)}=[p^{(i)\top},\tilde{s}^{(i)\top}]^{\top}\) have different
physical units and induce different local changes in the Gaussian. Directly
computing eigenvectors in this raw coordinate system can therefore make the
joint split direction sensitive to the relative parameterization of position
and scale.

To compare position and log-scale perturbations in a common coordinate system,
we use a local Gaussian-shape metric. Let
\[
\Sigma^{(i)}
=
L^{(i)}L^{(i)\top}
\]
be the covariance of the parent Gaussian, where
\(L^{(i)}=R^{(i)}\operatorname{diag}(s^{(i)})\) is the covariance factor aligned
with the principal scale axes. We measure infinitesimal perturbations of the
center and log-scale by
\begin{equation}
d_{\mathcal G}^{2}
=
\delta p^{(i)\top}
\Sigma^{(i)-1}
\delta p^{(i)}
+
2\|\delta \tilde{s}^{(i)}\|_{2}^{2}.
\label{eq:app_gaussian_local_metric}
\end{equation}
The first term is the Mahalanobis norm of the position displacement, so a unit
position perturbation corresponds to moving by one local Gaussian standard
deviation. The second term is a Fisher-inspired normalization for log-scale
perturbations of the underlying normalized Gaussian shape.

To see the origin of the factor \(2\), consider the one-dimensional normalized
Gaussian density
\[
q_{\mathcal N}(y;\mu,\tilde{s})
=
\frac{1}{\sqrt{2\pi}\exp(\tilde{s})}
\exp\left(
-\frac{1}{2}
\left(
\frac{y-\mu}{\exp(\tilde{s})}
\right)^2
\right),
\]
and let \(z=(y-\mu)/\exp(\tilde{s})\). Then
\[
\frac{\partial \log q_{\mathcal N}}{\partial \tilde{s}}
=
z^2-1.
\]
Since \(z\sim\mathcal N(0,1)\), \(\mathbb E[z^2]=1\) and
\(\mathbb E[z^4]=3\), giving
\begin{equation}
\mathbb E\left[
\left(
\frac{\partial \log q_{\mathcal N}}{\partial \tilde{s}}
\right)^2
\right]
=
\mathbb E[(z^2-1)^2]
=
2.
\label{eq:app_log_scale_fisher}
\end{equation}
Applying this independently along the three principal scale axes gives the
log-scale part of Eq.~\eqref{eq:app_gaussian_local_metric}.

We emphasize that this metric is used as a coordinate normalization for the
underlying Gaussian shape. The rendered 3DGS opacity contribution is an
unnormalized Gaussian kernel, so Eq.~\eqref{eq:app_gaussian_local_metric}
should not be interpreted as the exact Fisher metric of the rendered opacity
contribution. Instead, it provides a dimensionless local coordinate system in
which position and log-scale perturbations are compared on a common scale.

We introduce a normalized joint direction
\(v^{(i)}=[v_p^{(i)\top},v_{\tilde{s}}^{(i)\top}]^{\top}\in\mathbb R^{6}\)
and map it to a raw position-log-scale perturbation by
\begin{equation}
\delta \phi^{(i)}
=
\begin{bmatrix}
\delta p^{(i)}\\
\delta \tilde{s}^{(i)}
\end{bmatrix}
=
J^{(i)}v^{(i)}
=
\begin{bmatrix}
L^{(i)} & 0\\
0 & \frac{1}{\sqrt{2}}\mathbf I
\end{bmatrix}
\begin{bmatrix}
v_p^{(i)}\\
v_{\tilde{s}}^{(i)}
\end{bmatrix}
=
\begin{bmatrix}
L^{(i)}v_p^{(i)}\\
\frac{1}{\sqrt{2}}v_{\tilde{s}}^{(i)}
\end{bmatrix}.
\label{eq:app_joint_normalized_transform}
\end{equation}
Substituting these into Eq.~\eqref{eq:app_gaussian_local_metric} gives
\begin{equation}
d_{\mathcal G}^{2}
=
\|v_p^{(i)}\|_{2}^{2}
+
\|v_{\tilde{s}}^{(i)}\|_{2}^{2}
=
\|v^{(i)}\|_2^2.
\label{eq:app_normalized_metric}
\end{equation}
Thus, \(v^{(i)}\) is a dimensionless coordinate in which one unit of position
perturbation and one unit of log-scale perturbation are measured using the same
local Gaussian metric.

The raw second-order splitting surrogate for a perturbation
\(\delta\phi^{(i)}\) is
\[
\delta\phi^{(i)\top}
\bar{\mathcal S}_{\phi}^{(i)}
\delta\phi^{(i)}.
\]
Here, \(\bar{\mathcal S}_{\phi}^{(i)}\) denotes the symmetrized raw joint
splitting matrix used for eigendecomposition. Using
\(\delta\phi^{(i)}=J^{(i)}v^{(i)}\), this becomes
\begin{equation}
\delta\phi^{(i)\top}
\bar{\mathcal S}_{\phi}^{(i)}
\delta\phi^{(i)}
=
v^{(i)\top}
\left(
J^{(i)\top}
\bar{\mathcal S}_{\phi}^{(i)}
J^{(i)}
\right)
v^{(i)}.
\label{eq:app_normalized_surrogate}
\end{equation}
Therefore, the splitting matrix in the normalized joint coordinate system is
\begin{equation}
\tilde{\mathcal S}_{\phi}^{(i)}
=
J^{(i)\top}
\bar{\mathcal S}_{\phi}^{(i)}
J^{(i)}.
\label{eq:app_normalized_split_matrix}
\end{equation}
In practice, we symmetrize \(\mathcal S_{\phi}^{(i)}\) before eigendecomposition
to remove numerical asymmetry from accumulation. We then compute the split
direction as the eigenvector associated with the smallest eigenvalue of
\(\tilde{\mathcal S}_{\phi}^{(i)}\), and map this normalized direction back to
the raw position-log-scale perturbation using
Eq.~\eqref{eq:app_joint_normalized_transform}.
%%%%%%%%%%%%%%%%%%%%%%%%%%%%%%%%
\subsection{Proxy Curvature for Local Surrogates}
\label{sec:residual_curvature}

This section details the image-space curvature weight used in the clone and
opacity decay surrogates. In Sec.~\ref{sec:method_clone} and
Sec.~\ref{sec:method_opacity_decay}, both operations are modeled by a one-dimensional
quadratic approximation with respect to the operation strength. To define these
surrogates, we need a local curvature term that describes how the photometric
loss \(\mathcal L\) changes under an image-space perturbation. However, directly
computing the exact curvature of \(\mathcal L\) is challenging because
\(\mathcal L\) combines an \(\mathcal{L}_1\) term and an SSIM~\cite{ssim} term.
The \(\mathcal{L}_1\) component is non-smooth, while the SSIM component depends on
local-window statistics such as means, variances, and covariances, which couple
neighboring pixels. Therefore, instead of computing the full Hessian of
\(\mathcal L\), we construct a diagonal image-space curvature proxy that is
positive, stable, and inexpensive to accumulate during rasterization.

We first approximate the \(\mathcal{L}_1\) curvature using a Charbonnier robust
penalty~\cite{charbonnier1997deterministic,barron2019general}. Let the residual \(r_c(x)\)
at pixel \(x\) and color channel \(c\) be
\begin{equation}
r_c(x)
=
I_c(x)-\hat I_c(x),
\end{equation}
where \(I_c(x)\) and \(\hat I_c(x)\) denote the rendered and target color values, respectively, at pixel \(x\) in the \(c\)-th color channel.
The Charbonnier penalty is
\begin{equation}
\rho_{\epsilon}\!\left(r_c(x)\right)
=
\sqrt{r_c(x)^2+\epsilon^2}-\epsilon .
\end{equation}
where \(\epsilon\) is a small smoothing parameter controlling the transition
between quadratic behavior near zero residuals and \(\mathcal{L}_1\)-like behavior for
large residuals. We set \(\epsilon=10^{-3}\) in all experiments. We use the
curvature induced by this smooth penalty as the Charbonnier curvature weight:
\begin{equation}
\omega_{\mathrm{charb},c}(x)
=
\rho_{\epsilon}''\!\left(r_c(x)\right)
=
\frac{\epsilon^2}
{\left(r_c(x)^2+\epsilon^2\right)^{3/2}} .
\label{eq:charb_curvature_weight}
\end{equation}
This gives a smooth diagonal approximation to the \(\mathcal{L}_{1}\) curvature. To also account for the SSIM component, we follow recent 3DGS second-order optimizers that use diagonal approximations to handle its local pixel coupling efficiently~\cite{hollein20253dgs,pehlivan2026matrix}, and
introduce a diagonal SSIM curvature proxy. To approximate the SSIM curvature, we follow the
standard 3DGS SSIM setting and use an \(11\times 11\) Gaussian window with
standard deviation \(1.5\). Let \(G\) denote this normalized window, and let
\(*\) denote convolution over the image plane. For each color channel, we
compute target-window statistics:
\begin{equation}
\mu_c(x)
=
(G * \hat I_c)(x),
\qquad
\nu_c(x)
=
(G * \hat I_c^2)(x)-\mu_c(x)^2 ,
\end{equation}
where \(\mu_c(x)\) and \(\nu_c(x)\) denote the local target mean and variance.
We then form the inverse luminance and contrast denominators
\begin{equation}
D_{\mu,c}^{-1}(x)
=
\frac{1}
{2\mu_c(x)^2 + C_1 + \epsilon_{\mathrm{ssim}}},
\qquad
D_{\nu,c}^{-1}(x)
=
\frac{1}
{2\nu_c(x) + C_2 + \epsilon_{\mathrm{ssim}}}.
\end{equation}
Using these local denominators, we approximate the diagonal curvature induced by
the SSIM term as
\begin{equation}
\omega_{\mathrm{ssim},c}(x)
=
2\,(G^2 * D_{\mu,c}^{-1})(x)
+
2\left[
(G * D_{\nu,c}^{-1})(x)
-
(G^2 * D_{\nu,c}^{-1})(x)
\right],
\label{eq:ssim_diagonal_curvature}
\end{equation}
where \(G^2\) denotes the element-wise squared Gaussian window. This expression
keeps only the diagonal response of the local-window SSIM approximation and
drops cross-pixel Hessian terms, making it efficient to evaluate as separable
convolutions.

Finally, we combine the Charbonnier and SSIM curvature proxies according to the
same weighting used in \(\mathcal L\):
\begin{equation}
\omega_c(x)
=
(1-\lambda_{\mathrm{D\text{-}SSIM}})
\omega_{\mathrm{charb},c}(x)
+
\lambda_{\mathrm{D\text{-}SSIM}}
\gamma_{\mathrm{ssim}}
\omega_{\mathrm{ssim},c}(x).
\label{eq:mixed_proxy_curvature}
\end{equation}
where \(\gamma_{\mathrm{ssim}}\) controls the scale of the SSIM curvature proxy.  
We set \(\gamma_{\mathrm{ssim}}=10^{-4}\) empirically to balance the magnitude of
the SSIM proxy with the Charbonnier curvature. This mixed curvature weight is used as the proxy curvature in the clone and opacity decay surrogates. By combining the residual-based Charbonnier term
with the diagonal SSIM proxy, it provides a lightweight and effective
second-order approximation of the photometric loss \(\mathcal L\) under local
image-space perturbations. In practice, \(\omega_c(x)\) is computed once from the current
rendered image and target image, and then reused for all Gaussian candidates
contributing to the same pixel.

%%%%%%%%%%%%%%%%%%%%%%%%%%%%%%%%

\section{More Experiment Results}
\label{app:experiment}
\begin{table}[t]
\centering
\caption{PSNR scores for Mip-NeRF360, Tanks\&Temples, and Deep Blending scenes.}

\setlength{\tabcolsep}{3.5pt}
\renewcommand{\arraystretch}{1.1}
\resizebox{\linewidth}{!}{
\begin{tabular}{l|ccccccccc|cc|cc}
\toprule
 & \multicolumn{9}{c|}{Mip-NeRF360}
 & \multicolumn{2}{c|}{Tanks\&Temples}
 & \multicolumn{2}{c}{Deep Blending} \\
\cmidrule(lr){2-10} \cmidrule(lr){11-12} \cmidrule(lr){13-14}
 & bicycle & flowers & garden & stump & treehill & room & counter & kitchen & bonsai
 & train & truck & drjohnson & playroom \\
\midrule
3DGS    
& $25.19$ & $21.51$ & $27.31$ & $26.58$ & $22.31$ & $31.38$ & $28.93$ & $30.84$ & $32.05$ 
& $21.79$ & $24.99$ & $28.98$ & $30.07$ \\

AbsGS   
& $25.27_{\pm0.02}$ & $21.26_{\pm0.03}$ & $27.57_{\pm0.02}$ & $26.59_{\pm0.10}$ & $21.93_{\pm0.04}$ 
& $31.60_{\pm0.12}$ & $29.12_{\pm0.06}$ & $31.63_{\pm0.25}$ & $32.26_{\pm0.16}$ 
& $21.68_{\pm0.12}$ & $25.65_{\pm0.04}$ & $29.16_{\pm0.04}$ & $30.08_{\pm0.06}$ \\

PixelGS 
& $25.30_{\pm0.03}$ & $21.56_{\pm0.02}$ & $27.48_{\pm0.04}$ & $26.82_{\pm0.04}$ & $22.26_{\pm0.03}$ & $31.40_{\pm0.05}$ & $29.16_{\pm0.02}$ & $31.74_{\pm0.07}$ & $32.45_{\pm0.07}$ 
& $22.02_{\pm0.07}$ & $25.53_{\pm0.01}$ & $28.07_{\pm0.03}$ & $29.79_{\pm0.04}$ \\

DC4GS   
& $25.48_{\pm0.04}$ & $21.55_{\pm0.02}$ & $27.62_{\pm0.14}$ & $26.89_{\pm0.01}$ & $22.13_{\pm0.02}$ 
& $31.78_{\pm0.07}$ & $28.99_{\pm0.09}$ & $31.54_{\pm0.20}$ & $32.30_{\pm0.01}$ 
& $22.07_{\pm0.07}$ & $25.82_{\pm0.02}$ & $29.17_{\pm0.02}$ & $30.12_{\pm0.05}$ \\

GDAGS   
& $25.34_{\pm0.03}$ & $21.25_{\pm0.08}$ & $27.58_{\pm0.03}$ & $26.78_{\pm0.05}$ & $21.98_{\pm0.03}$ 
& $31.54_{\pm0.06}$ & $29.07_{\pm0.03}$ & $31.76_{\pm0.08}$ & $32.20_{\pm0.10}$ 
& $21.67_{\pm0.11}$ & $25.64_{\pm0.02}$ & $29.22_{\pm0.04}$ & $30.00_{\pm0.13}$ \\

ResGS   
& $25.28_{\pm0.02}$ & $21.66_{\pm0.04}$ & $27.59_{\pm0.01}$ & $26.67_{\pm0.03}$ & $22.77_{\pm0.06}$ & $31.85_{\pm0.08}$ & $29.02_{\pm0.02}$ & $31.69_{\pm0.06}$ & $32.34_{\pm0.04}$ 
& $22.23_{\pm0.08}$ & $25.81_{\pm0.03}$ & $29.21_{\pm0.20}$ & $29.87_{\pm0.31}$ \\

SteepGS 
& $24.81_{\pm0.01}$ & $20.72_{\pm0.03}$ & $27.14_{\pm0.02}$ & $25.87_{\pm0.05}$ & $22.18_{\pm0.09}$ 
& $31.35_{\pm0.09}$ & $28.73_{\pm0.02}$ & $30.86_{\pm0.17}$ & $31.80_{\pm0.03}$ 
& $21.74_{\pm0.09}$ & $25.22_{\pm0.01}$ & $29.29_{\pm0.05}$ & $29.99_{\pm0.06}$ \\

Ours    
& $25.31_{\pm0.03}$ & $21.72_{\pm0.01}$ & $27.59_{\pm0.07}$ & $26.62_{\pm0.02}$ & $22.83_{\pm0.06}$ & $32.07_{\pm0.08}$ & $29.13_{\pm0.01}$ & $31.78_{\pm0.02}$ & $32.46_{\pm0.03}$ 
& $22.44_{\pm0.03}$ & $25.96_{\pm0.02}$ & $29.44_{\pm0.06}$ & $30.14_{\pm0.14}$ \\
\bottomrule
\end{tabular}
}
\label{sup_psnr}
\end{table}

\begin{table}[t]
\centering
\caption{SSIM scores for Mip-NeRF360, Tanks\&Temples, and Deep Blending scenes.}

\setlength{\tabcolsep}{3.5pt}
\renewcommand{\arraystretch}{1.1}
\resizebox{\linewidth}{!}{
\begin{tabular}{l|ccccccccc|cc|cc}
\toprule
 & \multicolumn{9}{c|}{Mip-NeRF360}
 & \multicolumn{2}{c|}{Tanks\&Temples}
 & \multicolumn{2}{c}{Deep Blending} \\
\cmidrule(lr){2-10} \cmidrule(lr){11-12} \cmidrule(lr){13-14}
 & bicycle & flowers & garden & stump & treehill & room & counter & kitchen & bonsai
 & train & truck & drjohnson & playroom \\
\midrule

3DGS    
& $0.764$ & $0.605$ & $0.866$ & $0.771$ & $0.632$ 
& $0.918$ & $0.907$ & $0.925$ & $0.940$ 
& $0.808$ & $0.876$ & $0.899$ & $0.907$ \\

AbsGS   
& $0.782_{\pm0.00}$ & $0.621_{\pm0.00}$ & $0.872_{\pm0.00}$ & $0.777_{\pm0.00}$ & $0.614_{\pm0.00}$ 
& $0.925_{\pm0.00}$ & $0.912_{\pm0.00}$ & $0.928_{\pm0.00}$ & $0.945_{\pm0.00}$ 
& $0.817_{\pm0.00}$ & $0.887_{\pm0.00}$ & $0.896_{\pm0.00}$ & $0.906_{\pm0.00}$ \\

PixelGS 
& $0.779_{\pm0.00}$ & $0.636_{\pm0.00}$ & $0.871_{\pm0.00}$ & $0.785_{\pm0.00}$ & $0.634_{\pm0.00}$ 
& $0.922_{\pm0.00}$ & $0.915_{\pm0.00}$ & $0.931_{\pm0.00}$ & $0.946_{\pm0.00}$ 
& $0.826_{\pm0.00}$ & $0.887_{\pm0.00}$ & $0.888_{\pm0.00}$ & $0.905_{\pm0.00}$ \\

DC4GS   
& $0.792_{\pm0.00}$ & $0.635_{\pm0.00}$ & $0.874_{\pm0.00}$ & $0.793_{\pm0.00}$ & $0.631_{\pm0.00}$ & $0.925_{\pm0.00}$ & $0.911_{\pm0.00}$ & $0.929_{\pm0.00}$ & $0.945_{\pm0.00}$ 
& $0.825_{\pm0.00}$ & $0.889_{\pm0.00}$ & $0.902_{\pm0.00}$ & $0.910_{\pm0.00}$ \\

GDAGS   
& $0.786_{\pm0.00}$ & $0.626_{\pm0.00}$ & $0.872_{\pm0.00}$ & $0.785_{\pm0.00}$ & $0.622_{\pm0.00}$ 
& $0.924_{\pm0.00}$ & $0.911_{\pm0.00}$ & $0.930_{\pm0.00}$ & $0.944_{\pm0.00}$ 
& $0.818_{\pm0.00}$ & $0.886_{\pm0.00}$ & $0.900_{\pm0.00}$ & $0.907_{\pm0.00}$ \\

ResGS   
& $0.767_{\pm0.00}$ & $0.606_{\pm0.00}$ & $0.867_{\pm0.00}$ & $0.774_{\pm0.00}$ & $0.638_{\pm0.00}$ 
& $0.919_{\pm0.00}$ & $0.908_{\pm0.00}$ & $0.928_{\pm0.00}$ & $0.942_{\pm0.00}$ 
& $0.816_{\pm0.00}$ & $0.885_{\pm0.00}$ & $0.902_{\pm0.00}$ & $0.905_{\pm0.00}$ \\

SteepGS 
& $0.731_{\pm0.00}$ & $0.547_{\pm0.00}$ & $0.856_{\pm0.00}$ & $0.733_{\pm0.00}$ & $0.612_{\pm0.00}$ 
& $0.916_{\pm0.00}$ & $0.901_{\pm0.00}$ & $0.923_{\pm0.00}$ & $0.938_{\pm0.00}$ 
& $0.804_{\pm0.00}$ & $0.876_{\pm0.00}$ & $0.902_{\pm0.00}$ & $0.907_{\pm0.00}$ \\

Ours    
& $0.770_{\pm0.00}$ & $0.607_{\pm0.00}$ & $0.868_{\pm0.00}$ & $0.776_{\pm0.00}$ & $0.637_{\pm0.00}$ 
& $0.921_{\pm0.00}$ & $0.909_{\pm0.00}$ & $0.927_{\pm0.00}$ & $0.943_{\pm0.00}$ 
& $0.818_{\pm0.00}$ & $0.889_{\pm0.00}$ & $0.903_{\pm0.00}$ & $0.908_{\pm0.00}$ \\

\bottomrule
\end{tabular}
}
\label{sup_ssim}
\end{table}

\begin{table}[t]
\centering
\caption{LPIPS scores for Mip-NeRF360, Tanks\&Temples, and Deep Blending scenes.}

\setlength{\tabcolsep}{3.5pt}
\renewcommand{\arraystretch}{1.1}
\resizebox{\linewidth}{!}{
\begin{tabular}{l|ccccccccc|cc|cc}
\toprule
 & \multicolumn{9}{c|}{Mip-NeRF360}
 & \multicolumn{2}{c|}{Tanks\&Temples}
 & \multicolumn{2}{c}{Deep Blending} \\
\cmidrule(lr){2-10} \cmidrule(lr){11-12} \cmidrule(lr){13-14}
 & bicycle & flowers & garden & stump & treehill & room & counter & kitchen & bonsai
 & train & truck & drjohnson & playroom \\
\midrule

3DGS    
& $0.212$ & $0.336$ & $0.108$ & $0.216$ & $0.324$ 
& $0.220$ & $0.201$ & $0.128$ & $0.205$ 
& $0.214$ & $0.151$ & $0.245$ & $0.240$ \\

AbsGS   
& $0.172_{\pm0.00}$ & $0.271_{\pm0.00}$ & $0.099_{\pm0.00}$ & $0.197_{\pm0.00}$ & $0.281_{\pm0.00}$ 
& $0.200_{\pm0.00}$ & $0.188_{\pm0.00}$ & $0.122_{\pm0.00}$ & $0.190_{\pm0.00}$ 
& $0.193_{\pm0.00}$ & $0.132_{\pm0.00}$ & $0.241_{\pm0.00}$ & $0.232_{\pm0.00}$ \\

PixelGS 
& $0.180_{\pm0.00}$ & $0.261_{\pm0.00}$ & $0.099_{\pm0.00}$ & $0.188_{\pm0.00}$ & $0.277_{\pm0.00}$ 
& $0.210_{\pm0.00}$ & $0.183_{\pm0.00}$ & $0.119_{\pm0.00}$ & $0.191_{\pm0.00}$ 
& $0.179_{\pm0.00}$ & $0.120_{\pm0.00}$ & $0.255_{\pm0.00}$ & $0.241_{\pm0.00}$ \\

DC4GS   
& $0.166_{\pm0.00}$ & $0.269_{\pm0.00}$ & $0.099_{\pm0.00}$ & $0.188_{\pm0.00}$ & $0.272_{\pm0.00}$ 
& $0.202_{\pm0.00}$ & $0.190_{\pm0.00}$ & $0.122_{\pm0.00}$ & $0.191_{\pm0.00}$ 
& $0.189_{\pm0.00}$ & $0.132_{\pm0.00}$ & $0.238_{\pm0.00}$ & $0.233_{\pm0.00}$ \\

GDAGS   
& $0.171_{\pm0.00}$ & $0.274_{\pm0.00}$ & $0.099_{\pm0.00}$ & $0.193_{\pm0.00}$ & $0.279_{\pm0.00}$ 
& $0.204_{\pm0.00}$ & $0.192_{\pm0.00}$ & $0.122_{\pm0.00}$ & $0.192_{\pm0.00}$ 
& $0.197_{\pm0.00}$ & $0.137_{\pm0.00}$ & $0.239_{\pm0.00}$ & $0.233_{\pm0.00}$ \\

ResGS   
& $0.212_{\pm0.00}$ & $0.339_{\pm0.00}$ & $0.110_{\pm0.00}$ & $0.218_{\pm0.00}$ & $0.331_{\pm0.00}$ 
& $0.219_{\pm0.00}$ & $0.202_{\pm0.00}$ & $0.127_{\pm0.00}$ & $0.205_{\pm0.00}$ 
& $0.208_{\pm0.00}$ & $0.148_{\pm0.00}$ & $0.237_{\pm0.00}$ & $0.240_{\pm0.00}$ \\

SteepGS 
& $0.264_{\pm0.00}$ & $0.399_{\pm0.00}$ & $0.127_{\pm0.00}$ & $0.274_{\pm0.00}$ & $0.376_{\pm0.00}$ 
& $0.229_{\pm0.00}$ & $0.216_{\pm0.00}$ & $0.135_{\pm0.00}$ & $0.211_{\pm0.00}$ 
& $0.226_{\pm0.00}$ & $0.160_{\pm0.00}$ & $0.249_{\pm0.00}$ & $0.254_{\pm0.00}$ \\

Ours    
& $0.213_{\pm0.00}$ & $0.342_{\pm0.00}$ & $0.112_{\pm0.00}$ & $0.219_{\pm0.00}$ & $0.335_{\pm0.00}$ 
& $0.218_{\pm0.00}$ & $0.199_{\pm0.00}$ & $0.128_{\pm0.00}$ & $0.202_{\pm0.00}$ 
& $0.205_{\pm0.00}$ & $0.142_{\pm0.00}$ & $0.245_{\pm0.00}$ & $0.246_{\pm0.00}$ \\

\bottomrule
\end{tabular}
}
\label{sup_lpips}
\end{table}

\subsection{All scene results}
\label{app:per_scene}

We report the results for all scenes included in the main table in Tab.~\ref{sup_psnr}, \ref{sup_ssim}, and \ref{sup_lpips}. All results are computed based on the averages obtained from five different runs.

\subsection{Additional Ablation Study}
\label{app:ablation_additional}

We provide additional ablation results on Tanks\&Temples and Deep Blending
in Tab.~\ref{tab:ablation_additional}, using the same component
configurations as in Tab.~\ref{tab:ablation}.

\begin{table}[t]
\centering
\caption{
Ablation study of the proposed components on Tanks\&Temples and
Deep Blending. Variants (a)--(d) follow Tab.~\ref{tab:ablation}.
Num denotes the average number of Gaussians in millions.
}
\label{tab:ablation_additional}
\setlength{\tabcolsep}{2.5pt}
\renewcommand{\arraystretch}{1.1}
\resizebox{\linewidth}{!}{%
\begin{tabular}{c|cccc|cccc}
\toprule
Dataset
& \multicolumn{4}{c|}{Tanks\&Temples}
& \multicolumn{4}{c}{Deep Blending} \\
\cmidrule(lr){2-5} \cmidrule(lr){6-9}
Method
& PSNR$\uparrow$ & SSIM$\uparrow$ & LPIPS$\downarrow$ & Num
& PSNR$\uparrow$ & SSIM$\uparrow$ & LPIPS$\downarrow$ & Num \\
\midrule
(a): Baseline~\cite{steep}
& $23.48$ & $0.839$ & $0.193$ & $1.26$
& $29.59$ & $0.904$ & $0.251$ & $1.57$ \\
(b): (a) + Sec.~\ref{sec:method_split}
& $23.53$ & $0.842$ & $0.191$ & $1.27$
& $29.65$ & $0.907$ & $0.248$ & $1.64$ \\
(c): (b) + Sec.~\ref{sec:method_clone}
& $23.82$ & $0.845$ & $0.189$ & $\mathbf{0.98}$
& $29.70$ & $\mathbf{0.907}$ & $0.245$ & $\mathbf{1.31}$ \\
(d): (c) + Sec.~\ref{sec:method_opacity_decay}
& $\mathbf{24.20}$ & $\mathbf{0.853}$ & $\mathbf{0.174}$ & $1.04$
& $\mathbf{29.79}$ & $0.906$ & $\mathbf{0.245}$ & $1.32$ \\
\bottomrule
\end{tabular}%
}
\end{table}

\subsection{Matched-Budget Comparison}
\label{app:budget_dc4gs}

We compare our method with 3DGS~\cite{3dgs} and DC4GS~\cite{dc4gs}
under comparable Gaussian budgets, using the same scene setting as in the main experiments.
For DC4GS, we adjust the number of densification events to obtain
a final Gaussian count comparable to ours, while keeping its other
training settings unchanged.
As shown in Tab.~\ref{tab:budget_dc4gs}, our method achieves higher
PSNR and SSIM than DC4GS across all three benchmarks, whereas
DC4GS obtains lower LPIPS.

We additionally compare our method with 3DGS and
3DGS-MCMC~\cite{mcmc}, with the Gaussian counts of the baselines
matched to those of our method.
For Mip-NeRF360, we follow the scene selection of 3DGS-MCMC and
evaluate all compared methods on the seven publicly available scenes. As shown in Tab.~\ref{tab:budget_mcmc}, our method achieves the best
PSNR, SSIM, and LPIPS on Tanks\&Temples and Deep Blending.
On the seven-scene Mip-NeRF360 subset, our method achieves comparable
PSNR to 3DGS-MCMC, while 3DGS-MCMC obtains higher SSIM and lower LPIPS.

\begin{table}[t]
\centering
\caption{
Matched-budget comparison with 3DGS-MCMC.
Mip-NeRF360 results are averaged over seven publicly available scenes,
unlike the nine-scene results in Tab.~\ref{tab:budget_dc4gs}.
Num denotes the average number of Gaussians in millions.
}
\label{tab:budget_mcmc}
\setlength{\tabcolsep}{3pt}
\renewcommand{\arraystretch}{1.1}
\resizebox{\linewidth}{!}{%
\begin{tabular}{c|cccc|cccc|cccc}
\toprule
Dataset
& \multicolumn{4}{c|}{Mip-NeRF360 (7 scenes)}
& \multicolumn{4}{c|}{Tanks\&Temples}
& \multicolumn{4}{c}{Deep Blending} \\
\cmidrule(lr){2-5} \cmidrule(lr){6-9} \cmidrule(lr){10-13}
Method
& PSNR$\uparrow$ & SSIM$\uparrow$ & LPIPS$\downarrow$ & Num
& PSNR$\uparrow$ & SSIM$\uparrow$ & LPIPS$\downarrow$ & Num
& PSNR$\uparrow$ & SSIM$\uparrow$ & LPIPS$\downarrow$ & Num \\
\midrule
3DGS~\cite{3dgs}
& $28.90$ & $0.863$ & $0.196$ & $1.8$
& $23.62$ & $0.842$ & $0.186$ & $1.0$
& $29.48$ & $0.901$ & $0.254$ & $1.3$ \\
3DGS-MCMC~\cite{mcmc}
& $29.28$ & $\mathbf{0.885}$ & $\mathbf{0.168}$ & $1.8$
& $24.04$ & $0.852$ & $0.176$ & $1.0$
& $29.53$ & $0.897$ & $0.266$ & $1.3$ \\
Ours
& $\mathbf{29.30}$ & $0.874$ & $0.184$ & $1.8$
& $\mathbf{24.20}$ & $\mathbf{0.853}$ & $\mathbf{0.174}$ & $1.0$
& $\mathbf{29.79}$ & $\mathbf{0.906}$ & $\mathbf{0.245}$ & $1.3$ \\
\bottomrule
\end{tabular}%
}
\end{table}

\begin{table}[t]
\centering
\caption{
Comparison with DC4GS under comparable Gaussian budgets.
}
\label{tab:budget_dc4gs}
\setlength{\tabcolsep}{3pt}
\renewcommand{\arraystretch}{1.1}
\resizebox{\linewidth}{!}{%
\begin{tabular}{c|cccc|cccc|cccc}
\toprule
Dataset
& \multicolumn{4}{c|}{Mip-NeRF360 (9 scenes)}
& \multicolumn{4}{c|}{Tanks\&Temples}
& \multicolumn{4}{c}{Deep Blending} \\
\cmidrule(lr){2-5} \cmidrule(lr){6-9} \cmidrule(lr){10-13}
Method
& PSNR$\uparrow$ & SSIM$\uparrow$ & LPIPS$\downarrow$ & Num
& PSNR$\uparrow$ & SSIM$\uparrow$ & LPIPS$\downarrow$ & Num
& PSNR$\uparrow$ & SSIM$\uparrow$ & LPIPS$\downarrow$ & Num \\
\midrule
3DGS~\cite{3dgs}
& $27.32$ & $0.804$ & $0.233$ & $1.9$
& $23.62$ & $0.842$ & $0.186$ & $1.0$
& $29.48$ & $0.901$ & $0.254$ & $1.3$ \\
DC4GS~\cite{dc4gs}
& $27.52$ & $0.816$ & $\mathbf{0.201}$ & $1.9$
& $23.95$ & $0.852$ & $\mathbf{0.168}$ & $1.0$
& $29.65$ & $0.901$ & $\mathbf{0.244}$ & $1.3$ \\
Ours
& $\mathbf{27.72}$ & $\mathbf{0.818}$ & $0.219$ & $1.9$
& $\mathbf{24.20}$ & $\mathbf{0.853}$ & $0.174$ & $1.0$
& $\mathbf{29.79}$ & $\mathbf{0.906}$ & $0.245$ & $1.3$ \\
\bottomrule
\end{tabular}%
}
\end{table}
\subsection{Convergence Analysis}
\label{app:convergence}

\begin{figure}[t]
    \centering
    \includegraphics[width=0.9\linewidth]{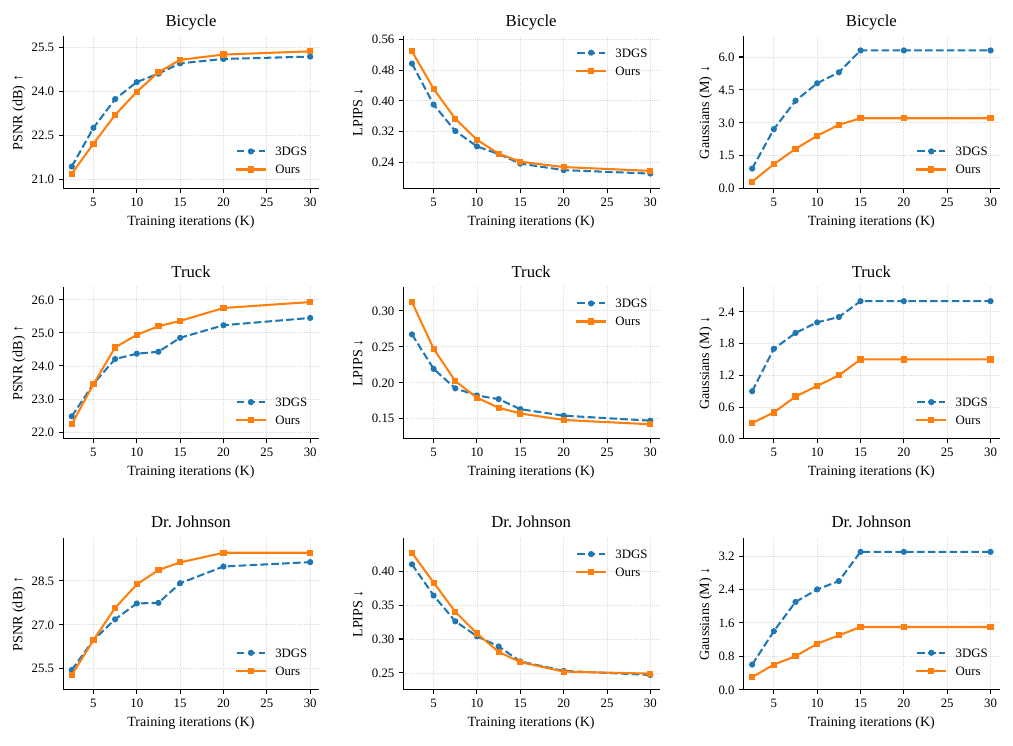}
    \caption{
    Convergence comparison with 3DGS on Bicycle, Truck, and
    Dr.~Johnson. Columns show PSNR, LPIPS, and the number of
    Gaussians.
    }
    \label{fig:convergence}
\end{figure}

We compare the convergence behavior of our method with
3DGS~\cite{3dgs} by tracking PSNR, LPIPS, and the number of
Gaussians over 30K training iterations.
Following the representative-scene evaluation setting of
SAD-GS~\cite{sadgs}, we use Bicycle from Mip-NeRF360,
Truck from Tanks\&Temples, and Dr.~Johnson from Deep Blending.
Fig.~\ref{fig:convergence} presents the results for each scene.

Our method initially uses substantially fewer Gaussians with
slightly lower reconstruction quality.
As training progresses, it surpasses 3DGS in PSNR at the
12.5K checkpoint on Bicycle and the 7.5K checkpoint on Truck
and Dr.~Johnson.
At 20K iterations, our method already exceeds the PSNR achieved
by 3DGS at 30K iterations on all three scenes, while using
substantially fewer Gaussians.
These results show that our loss-aware density control improves
mid-to-late-stage PSNR at a given iteration budget while maintaining
a more compact Gaussian representation.

The LPIPS trends are scene-dependent.
At 30K iterations, our method achieves lower LPIPS on Truck,
whereas 3DGS obtains slightly lower LPIPS on Bicycle and Dr.~Johnson. 

\subsection{Regional Analysis and Failure Cases}
\label{app:regional_analysis}

To examine where our method performs well or underperforms, we
conduct a patch-level comparison with DC4GS~\cite{dc4gs}.
We divide each rendered image into an $8 \times 10$ grid and
compute PSNR, SSIM, and LPIPS for each patch.
For each metric, we inspect the top-50 patches showing the largest
improvements and the top-50 showing the largest degradations
relative to DC4GS across all evaluated datasets, scenes, and
test images.

\begin{figure}[t]
    \centering
    \includegraphics[width=\linewidth]{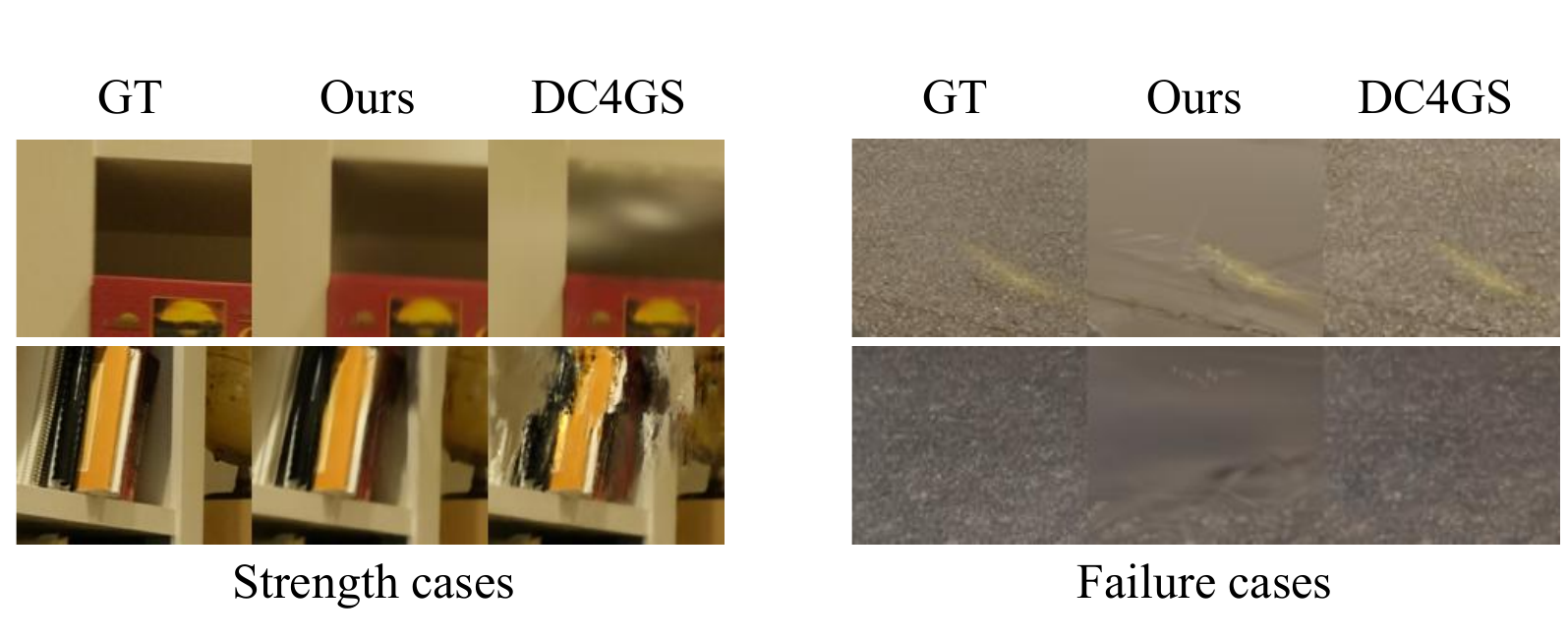}
    \caption{
    Patch-level comparison with DC4GS.
    The examples illustrate strengths in reconstructing object
    structures and object-background boundaries, as well as
    limitations in preserving fine details in repeated
    high-frequency textures.
    }
    \label{fig:regional_analysis}
\end{figure}

Among the inspected patches, our method effectively reconstructs
object structures and object-background boundaries.
However, preserving fine-grained details in regions with repeated
high-frequency textures, such as roads, grass, gravel, and foliage,
remains challenging.
In these regions, our reconstructions tend to appear smoother
than those of DC4GS.
Fig.~\ref{fig:regional_analysis} presents representative examples
of these strengths and limitations.

A possible explanation lies in our loss-aware allocation criterion,
which evaluates density-control operations according to their
predicted local reduction in photometric loss.
In texture-rich regions, a few relatively large Gaussians may
already approximate the dominant colors and low-frequency
appearance, leaving only small predicted gains from further
splitting or cloning.
Consequently, the method may allocate insufficient additional
Gaussians to preserve fine textures.
This behavior may partially explain the LPIPS gap with DC4GS,
which persists even under comparable Gaussian budgets
(Tab.~\ref{tab:budget_dc4gs}). We therefore consider improving perceptual fidelity in high-frequency
texture regions within our loss-aware framework an important direction
for future work.

\section{Dataset licenses}
\label{app:license}
\begin{itemize}[leftmargin=1em]

    \item Mip-NeRF 360~\cite{mipnerf360}: no license terms provided.
    Available at \url{https://jonbarron.info/mipnerf360/}.

    \item Deep Blending~\cite{Deepblending}: no license terms provided.
    Available at \url{http://visual.cs.ucl.ac.uk/pubs/deepblending/}.

    \item Tanks \& Temples~\cite{tankandtemples}: made available under
    Creative Commons Attribution-NonCommercial-ShareAlike 3.0 License.
    \url{https://www.tanksandtemples.org/license/}.
\end{itemize}

\section{Broader Impact}
\label{broader_impact}
Our work improves the efficiency of 3D Gaussian Splatting by replacing heuristic density-control operations with loss-aware, primitive-wise decisions. By reducing redundant Gaussian allocation and avoiding abrupt global opacity reset, the proposed method can lower memory and rendering costs while maintaining reconstruction quality. This may support more practical deployment of real-time 3D scene representations on resource-constrained devices, including mobile platforms and AR/VR systems, and may reduce storage, bandwidth, and computational demands.

However, more efficient 3D reconstruction can also make it easier to capture and distribute detailed representations of real-world environments. Such capability may raise privacy concerns when applied to private spaces, people, or sensitive objects without consent, and could be used in surveillance or synthetic media pipelines. 

% \section{Limitations and Future Work.}
% \label{limitations}
% Although our method provides a loss-aware density control strategy for 3D Gaussian Splatting, it has ㅅ limitations. 
% First, as described in Sec.~\ref{sec:residual_curvature}, our curvature estimate is an approximation of the full training objective. This approximation may introduce a discrepancy between the estimated curvature and potentially leading to suboptimal densification decisions.
% Second, our surrogate approximates each operation point-wise. While rendering interactions are implicitly reflected in the back-propagated signals, it does not explicitly model cross-Gaussian effects.

% Future directions involve curvature estimation that more faithfully reflects the full training objective, split operations that consider parameters beyond position and scale, and interaction-aware surrogate modeling for more reliable density control.

\section{Limitations and Future Work.}
\label{limitations}

While our method provides a loss-aware density control strategy for 3D Gaussian Splatting, it relies on tractable local surrogate approximations to enable efficient per-Gaussian decisions. This design choice suggests several directions for further refinement.
First, as described in Sec.~\ref{sec:residual_curvature}, our curvature estimate is an approximation of the full training objective. This approximation may introduce a discrepancy between the estimated curvature and the true objective curvature, which can occasionally lead to suboptimal densification decisions.
Second, our surrogate approximates each operation point-wise. While rendering interactions are implicitly reflected in the back-propagated signals, it does not explicitly model cross-Gaussian effects. Third, the additional surrogate computations increase training time and peak GPU memory usage, although they are not required at inference (Tab.~\ref{tab:computational_cost}).
Finally, preserving fine details in repeated high-frequency textures
remains challenging, which can limit perceptual fidelity
(Sec.~\ref{app:regional_analysis}).

Future directions involve curvature estimation that more faithfully
reflects the full training objective, split operations that consider parameters beyond position and scale, interaction-aware surrogate modeling, and improvements in training efficiency and perceptual fidelity within our loss-aware framework.

%%%%%%%%%%%%%%%%%%%%%%%%%%%%%%%%%%%%%%%%%%%%%%%%%%%%%%%%%%%%
\clearpage

\section*{NeurIPS Paper Checklist}

\begin{enumerate}

\item {\bf Claims}
    \item[] Question: Do the main claims made in the abstract and introduction accurately reflect the paper's contributions and scope?
    \item[] Answer: \answerYes{}{} % Replace by \answerYes{}, \answerNo{}, or \answerNA{}.
    \item[] Justification: The claims in the abstract and introduction accurately reflect the proposed method and are supported by the theoretical derivations and experimental results.

\item {\bf Limitations}
    \item[] Question: Does the paper discuss the limitations of the work performed by the authors?
    \item[] Answer: \answerYes{} % Replace by \answerYes{}, \answerNo{}, or \answerNA{}.
    \item[] Justification: We discuss the limitations of our method in Appendix~\ref{limitations}.

\item {\bf Theory assumptions and proofs}
    \item[] Question: For each theoretical result, does the paper provide the full set of assumptions and a complete (and correct) proof?
    \item[] Answer: \answerYes{} % Replace by \answerYes{}, \answerNo{}, or \answerNA{}.
    \item[] Justification: We state the assumptions for our theoretical derivations and provide the corresponding proofs and derivations in the Appendix.

    \item {\bf Experimental result reproducibility}
    \item[] Question: Does the paper fully disclose all the information needed to reproduce the main experimental results of the paper to the extent that it affects the main claims and/or conclusions of the paper (regardless of whether the code and data are provided or not)?
    \item[] Answer: \answerYes{} % Replace by \answerYes{}, \answerNo{}, or \answerNA{}.
    \item[] Justification: We provide the implementation details in Sec.~\ref{implementation_detail} and Appendix~\ref{app:hyperparameters}, and include the implementation code in the supplementary material.

\item {\bf Open access to data and code}
    \item[] Question: Does the paper provide open access to the data and code, with sufficient instructions to faithfully reproduce the main experimental results, as described in supplemental material?
    \item[] Answer: \answerYes{} % Replace by \answerYes{}, \answerNo{}, or \answerNA{}.
    \item[] Justification: We attach the implementation code in supplementary.

\item {\bf Experimental setting/details}
    \item[] Question: Does the paper specify all the training and test details (e.g., data splits, hyperparameters, how they were chosen, type of optimizer) necessary to understand the results?
    \item[] Answer: \answerYes{} % Replace by \answerYes{}, \answerNo{}, or \answerNA{}.
    \item[] Justification: We report the experiment details used in our experiments in Sec.~\ref{implementation_detail} and Appendix.~\ref{app:hyperparameters}.

\item {\bf Experiment statistical significance}
    \item[] Question: Does the paper report error bars suitably and correctly defined or other appropriate information about the statistical significance of the experiments?
    \item[] Answer: \answerYes{} % Replace by \answerYes{}, \answerNo{}, or \answerNA{}.
    \item[] Justification: We report the mean and standard deviation of each evaluation metric over five runs in Tab.~\ref{tab:main_results} and Appendix.~\ref{app:experiment}.

\item {\bf Experiments compute resources}
    \item[] Question: For each experiment, does the paper provide sufficient information on the computer resources (type of compute workers, memory, time of execution) needed to reproduce the experiments?
    \item[] Answer: \answerYes{} % Replace by \answerYes{}, \answerNo{}, or \answerNA{}.
    \item[] Justification: We report the computational resources used in our experiments in Sec.~\ref{implementation_detail}.
    
\item {\bf Code of ethics}
    \item[] Question: Does the research conducted in the paper conform, in every respect, with the NeurIPS Code of Ethics \url{https://neurips.cc/public/EthicsGuidelines}?
    \item[] Answer: \answerYes{} % Replace by \answerYes{}, \answerNo{}, or \answerNA{}.
    \item[] Justification: We have reviewed the NeurIPS Code of Ethics.

\item {\bf Broader impacts}
    \item[] Question: Does the paper discuss both potential positive societal impacts and negative societal impacts of the work performed?
    \item[] Answer: \answerYes{} % Replace by \answerYes{}, \answerNo{}, or \answerNA{}.
    \item[] Justification: We provide this in Appendix.~\ref{broader_impact}.
    
\item {\bf Safeguards}
    \item[] Question: Does the paper describe safeguards that have been put in place for responsible release of data or models that have a high risk for misuse (e.g., pre-trained language models, image generators, or scraped datasets)?
    \item[] Answer: \answerNA{} % Replace by \answerYes{}, \answerNo{}, or \answerNA{}.
    \item[] Justification: This work does not involve personally identifiable information, private user data, or sensitive datasets, and therefore does not introduce direct privacy-related risks.

\item {\bf Licenses for existing assets}
    \item[] Question: Are the creators or original owners of assets (e.g., code, data, models), used in the paper, properly credited and are the license and terms of use explicitly mentioned and properly respected?
    \item[] Answer: \answerYes{} % Replace by \answerYes{}, \answerNo{}, or \answerNA{}.
    \item[] Justification: We provide the licenses of the datasets used in our experiments in Appendix~\ref{app:license}.

\item {\bf New assets}
    \item[] Question: Are new assets introduced in the paper well documented and is the documentation provided alongside the assets?
    \item[] Answer: \answerNA{} % Replace by \answerYes{}, \answerNo{}, or \answerNA{}.
    \item[] Justification: We do not provide new assets.

\item {\bf Crowdsourcing and research with human subjects}
    \item[] Question: For crowdsourcing experiments and research with human subjects, does the paper include the full text of instructions given to participants and screenshots, if applicable, as well as details about compensation (if any)? 
    \item[] Answer: \answerNA{} % Replace by \answerYes{}, \answerNo{}, or \answerNA{}.
    \item[] Justification: This work does not include crowdsourcing experiments or studies involving human participants.

\item {\bf Institutional review board (IRB) approvals or equivalent for research with human subjects}
    \item[] Question: Does the paper describe potential risks incurred by study participants, whether such risks were disclosed to the subjects, and whether Institutional Review Board (IRB) approvals (or an equivalent approval/review based on the requirements of your country or institution) were obtained?
    \item[] Answer: \answerNA{} % Replace by \answerYes{}, \answerNo{}, or \answerNA{}.
    \item[] Justification: This work does not include crowdsourcing experiments or studies involving human participants.

\item {\bf Declaration of LLM usage}
    \item[] Question: Does the paper describe the usage of LLMs if it is an important, original, or non-standard component of the core methods in this research? Note that if the LLM is used only for writing, editing, or formatting purposes and does \emph{not} impact the core methodology, scientific rigor, or originality of the research, declaration is not required.
    %this research? 
    \item[] Answer: \answerNA{} % Replace by \answerYes{}, \answerNo{}, or \answerNA{}.
    \item[] Justification: We used LLMs only for auxiliary purposes, such as polishing English sentences, editing, and LaTeX formatting.

\end{enumerate}

\end{document}